\pdfoutput=1
\documentclass{article}

\PassOptionsToPackage{numbers,sort&compress}{natbib}
\usepackage[main, final]{neurips_2026}

\usepackage[utf8]{inputenc}
\usepackage[T1]{fontenc}
\usepackage{hyperref}
\hypersetup{
  pdftitle={OASIS: Observation-Action Space Alignment via SE(3) Trajectory Prediction for Robotic Manipulation},
  pdfauthor={Xinzhe Chen, Sihua Ren, Liqi Huang, Haowen Sun, Mingyang Li, Zeyang Liu, Xingyu Chen, Xuguang Lan}
}
\usepackage{url}
\usepackage{booktabs}
\usepackage{amsfonts}
\usepackage{nicefrac}
\usepackage{microtype}
\usepackage{xcolor}
\definecolor{markgreen}{HTML}{1A9641}
\definecolor{markred}{HTML}{D7191C}
\newcommand{\yesmark}{\textcolor{markred}{$\checkmark$}}
\newcommand{\nomark}{\textcolor{markgreen}{$\times$}}
\definecolor{oursrow}{gray}{0.95}
\newlength{\oursbandgap}
\newcommand{\oursbandheight}{\dimexpr\belowrulesep+\ht\strutbox+\dp\strutbox+\aboverulesep-2\oursbandgap\relax}
\newcommand{\oursmidrule}{\specialrule{\lightrulewidth}{\aboverulesep}{0pt}%
  \noalign{\vskip\oursbandgap{\color{oursrow}\hrule height \oursbandheight}%
  \vskip-\dimexpr\oursbandheight+\oursbandgap\relax\vskip\belowrulesep}}
\newcommand{\oursbottomrule}{\noalign{\vskip\aboverulesep}\specialrule{\heavyrulewidth}{0pt}{\belowbottomsep}}

\usepackage{graphicx}
\usepackage{subcaption}
\usepackage{amsmath}
\usepackage{amssymb}
\usepackage{mathtools}
\usepackage{amsthm}
\usepackage[capitalize,noabbrev]{cleveref}
\usepackage{makecell}
\usepackage{multirow}
\usepackage{wrapfig}
\usepackage[disable,textsize=tiny]{todonotes}
\usepackage{pgfplots}
\pgfplotsset{compat=1.18}
\usetikzlibrary{patterns}
\renewcommand{\acksection}{\section*{Acknowledgments}}

\title{OASIS: Observation-Action Space Alignment via SE(3) Trajectory Prediction for Robotic Manipulation}

\author{%
  Xinzhe Chen\thanks{Equal contribution.} \quad
  Sihua Ren\footnotemark[1] \quad
  Liqi Huang \quad
  Haowen Sun \quad
  Mingyang Li \\
  \bf Zeyang Liu \quad
  Xingyu Chen\footnotemark[2] \quad
  Xuguang Lan\thanks{Corresponding authors: \texttt{xglan@mail.xjtu.edu.cn} and \texttt{chenxingyu1990@xjtu.edu.cn}.} \\
  \mdseries National Key Laboratory of Human-Machine Hybrid Augmented Intelligence \\
  Institute of Artificial Intelligence and Robotics \\
  Xi'an Jiaotong University, Xi'an, China
}

\begin{document}

\maketitle

\begin{abstract}
Recent vision-language-action (VLA) models and world action models (WAMs)
advance robotic manipulation by enriching intermediate representations
with auxiliary spatial features or future visual-state prediction.
However, these representations largely remain within the observation
space and do not share the rigid-body geometry of the action space,
forcing the action decoder to implicitly recover this geometry. We
propose OASIS, a visuomotor policy that aligns the intermediate
representation with the action space via $SE(3)$ end-effector trajectory
prediction. OASIS couples a 3D-aware feature encoder that fuses
vision-language and metric-depth features with an $SE(3)$ trajectory
predictor that produces a camera-frame end-effector trajectory.
Conditioned on the predictor's pose-supervised hidden states, the action
decoder generates action chunks consistent with rigid-body motion.
Across simulation and real-world experiments, OASIS outperforms VLA and
WAM baselines in success rate and out-of-distribution generalization.
Our project page is available at \url{https://oasis-neurips.github.io}.
\end{abstract}

\section{Introduction}
\label{sec:introduction}

Visuomotor policies for robotic manipulation use neural networks to map image observations,
language instructions, and the robot state directly to executable 6-DoF actions and gripper
commands~\cite{rt1, gr2, rtx, rt2, lin2025data, zhang2024oneshot}, and three representative families of methods define the current frontier.
Vanilla Vision-Language-Action (VLA) models~\cite{bjorck2025gr00t, li2024visionlanguage, li2023generalist} build multimodal features from image
observations and language instructions, typically via a pretrained Vision-Language Model
(VLM), and feed these features directly into an action
decoder~\cite{pmlr-v270-kim25c, pmlr-v305-black25a}. Feature-enhanced VLAs~\cite{ranasinghe2025pixel, wen2023any, jia2024lift3d, qu2025spatialvla} strengthen
spatial understanding by injecting spatial features such as depth maps~\cite{li2025qdepth, zhen20243d},
regions of interest~\cite{song2026reconvla},
or 2D motion trajectories on the image plane~\cite{huang2025thinkact, lee2025molmoact}. World Action Models (WAMs)~\cite{du2023learning, xie2025latent, cen2025worldvla} take a different route,
learning a world model that predicts future images~\cite{hu2025video,
 zhang2025dreamvla} or latent visual features~\cite{liu-niu2025lbp, tian2025predictive,wang2025unified}.

\begin{figure}
  \centering
  \includegraphics[width=\columnwidth]{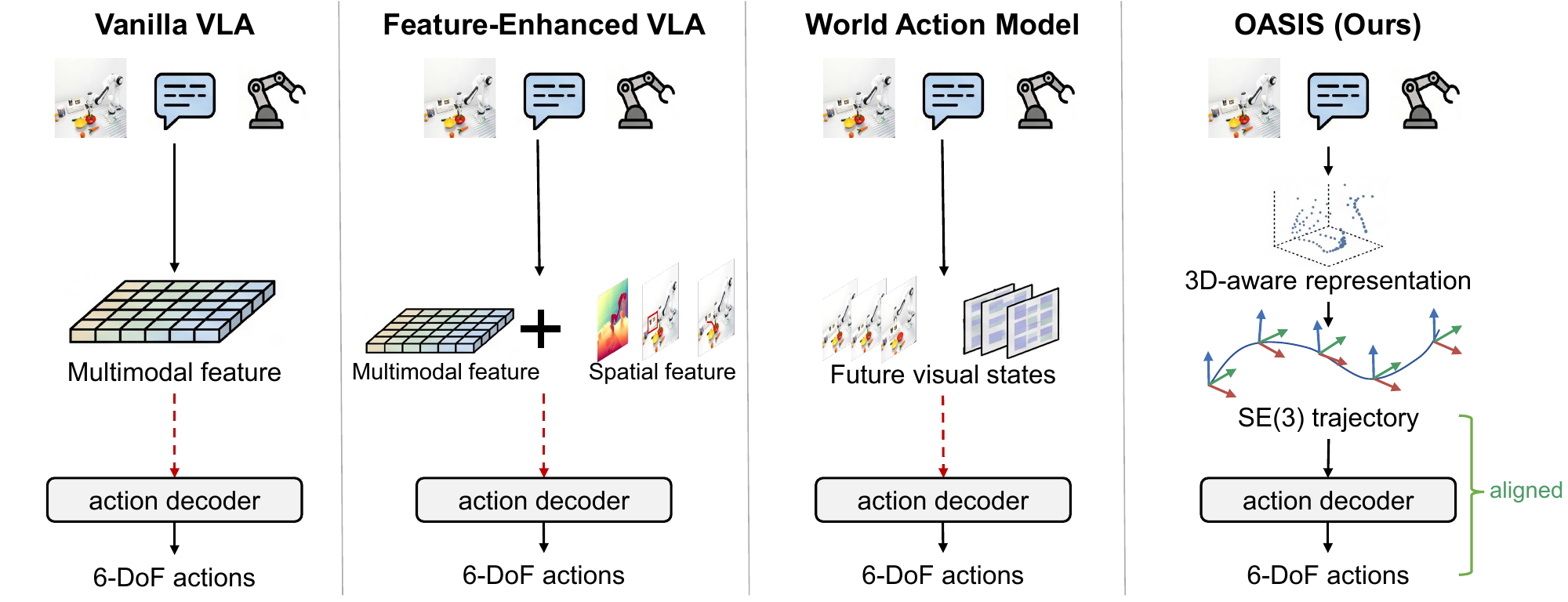}
    \caption{Comparison of existing visuomotor policies and OASIS.
    VLA models and WAMs construct intermediate representations
    that are not aligned with the action space.
    OASIS instead aligns the intermediate representation with the action space via $SE(3)$ trajectory prediction.}
  \label{fig:problem}
  \vspace{-8pt}
\end{figure}

Despite their empirical strength, these families decode 6-DoF relative actions
from intermediate representations that do not share the rigid-body geometry of
the action space. This mismatch matters because rigid-body motions of a robot
end-effector live in the Special Euclidean group $SE(3)$, yet a single image
observation is consistent with many distinct 6-DoF actions under the same task
instruction. Resolving this ambiguity requires an intermediate representation
that constrains the target $SE(3)$ pose. Large-scale robotic pretraining or auxiliary spatial supervision can mitigate this ambiguity,
but neither injects the geometric structure of $SE(3)$ into the intermediate.

\textbf{Our key insight is that predicting an $SE(3)$ end-effector
trajectory aligns the intermediate representation with the action space,
providing a geometric inductive bias rather than enriching
observation-space features with auxiliary spatial cues or predictive
visual capacity.}
Figure~\ref{fig:problem} contrasts this design with existing intermediates, and Section~\ref{sec:geometric_inconsistency} formalizes it.
An intermediate representation is geometrically aligned when it
provides an explicit readout of the target $SE(3)$ pose at each future step. Auxiliary supervision via future images,
latent visual features, or 2D trajectories leaves pose recovery implicit
in the decoder. In contrast, $SE(3)$ end-effector trajectory prediction
supervises the intermediate against the rigid-body geometry of the
action space, so the decoder receives explicit pose information for action generation.

We instantiate this insight as \textbf{OASIS} (\textbf{O}bservation-\textbf{A}ction
\textbf{S}pace Al\textbf{I}gnment via \textbf{S}E(3) Trajectory Prediction),
which couples three components trained end-to-end. A 3D-aware feature
encoder fuses vision-language and metric-depth features into a 3D-aware representation.
On this representation, an $SE(3)$ trajectory predictor forecasts a sequence of end-effector poses
in the camera frame. Sharing the frame of the representation spares the predictor
from learning the camera-to-world extrinsic.
The extrinsic is used only to build camera-frame training labels, so OASIS needs no extrinsic input at inference.
The action decoder attends to the predictor's pose-supervised hidden states and the
current end-effector state, and generates the action chunk while absorbing residuals
such as the frame conversion and the gripper-timing offset. Supervision comes
solely from standard expert demonstrations, so OASIS requires neither annotated spatial
labels nor pretraining on large-scale robotic demonstrations, and uses only generically
pretrained backbones.

Extensive experiments show that OASIS improves performance both on
simulation benchmarks and on real robots. OASIS achieves a \textbf{97.6\%} average success rate
on LIBERO~\cite{liu2023libero} and a \textbf{4.57} average sequence length on
CALVIN ABC$\to$D~\cite{mees2022calvin}. On two real robots, it reaches an 89.2\% average success rate,
outperforming $\pi_{0.5}$~\cite{pmlr-v305-black25a}, RDT~\cite{liu2025rdtb}, Seer-Large~\cite{tian2025predictive}, and ACT~\cite{zhao2023learning}.
Under out-of-distribution (OOD) perturbations in multi-task settings,
covering unseen backgrounds, a displaced third-person camera, and human interference,
OASIS attains a 90.8\% average success rate and loses less than $\pi_{0.5}$ under each perturbation.
Adding the $SE(3)$ trajectory predictor to the encoder and decoder raises success
from 89.5\% to 95.2\% on LIBERO-Long, the largest \emph{additional} gain among ablated components.
Our contributions are summarized as follows.

\begin{itemize}
    \item We formulate an aligned-intermediate design principle for visuomotor policies, identifying $SE(3)$ end-effector trajectory prediction as the mechanism that aligns the intermediate representation with the action space, providing a geometric inductive bias for action generation.

    \item We construct OASIS, a visuomotor policy coupling a 3D-aware feature encoder, an $SE(3)$ trajectory predictor, and an action decoder, trained end-to-end on standard expert demonstrations without spatial annotations or large-scale robotic pretraining.

    \item Across simulation and real-world settings, OASIS achieves the highest average performance among the compared methods and degrades less than $\pi_{0.5}$ under OOD perturbations, with ablations attributing the largest additional gain to the $SE(3)$ trajectory predictor.
\end{itemize}

\section{Related Work}
\label{sec:related}

\paragraph{Vanilla Vision-Language-Action Models.} Vanilla VLA models map image
observations and language instructions directly to 6-DoF actions through a multimodal
encoder and a dedicated action decoder~\cite{yue2024deervla,bu2024towards}.
Early work generates action chunks with autoregressive~\cite{sun2023smart} or denoising action heads~\cite{octo_2023}
such as ACT~\cite{zhao2023learning} and Diffusion Policy~\cite{chi2025diffusion}, followed
by transformer-based variants~\cite{Reuss-RSS-24,reuss2025efficient}
including RDT-1B~\cite{liu2025rdtb}. More recent methods leverage pretrained
vision-language models~\cite{li2024visionlanguage, bjorck2025gr00t} for stronger semantic priors, including OpenVLA~\cite{pmlr-v270-kim25c}
on a Llama VLM~\cite{touvron2023llama} and $\pi_{0}$~\cite{BlackK-RSS-25, pmlr-v305-black25a}
on a PaliGemma VLM~\cite{beyer2024paligemma} with flow matching~\cite{lipman2023flow}.
The action decoder must recover depth and rotation from multimodal features alone,
without a geometrically aligned intermediate.

\paragraph{Feature-Enhanced VLA Models.} Feature-enhanced VLA models improve this
image-to-action mapping by injecting auxiliary spatial features.
One line lifts observations into 3D space~\cite{zhen20243d, li2025pointvla, Ze2024DP3,jia2024lift3d,li2025bridgevla}.
SpatialVLA~\cite{qu2025spatialvla} encodes 3D position information via Ego3D position
encoding, and QDepth-VLA~\cite{li2025qdepth} introduces quantized depth as auxiliary
supervision. A complementary line provides task-relevant visual guidance~\cite{zheng2025tracevla,lee2025molmoact,wen2023any}.
ReconVLA~\cite{song2026reconvla} reconstructs gaze regions to focus attention on
target objects, ThinkAct~\cite{huang2025thinkact} predicts 2D end-effector
trajectories, and UniVLA~\cite{BuQ-RSS-25} derives task-centric action
representations from videos. These features strengthen spatial reasoning and task
understanding, yet the resulting intermediates do not share the rigid-body geometry
of the action space, leaving the decoder to bridge them to actions. Structured-pose
policies~\cite{shridhar2023perceiver, goyal2024rvt, gervet2023act3d, ke20243d}
instead treat $SE(3)$ poses as the final action representation, predicted from 3D
inputs that require additional sensors or multi-view reconstruction. OASIS, in
contrast, predicts an $SE(3)$ trajectory from RGB images as the intermediate that
conditions a learned action decoder.

\paragraph{World Action Models.} World Action Models predict future images or
latent visual features and decode actions from these predicted
targets~\cite{xie2025latent, zhang2024pivot}. SuSIE~\cite{black2023zero} synthesizes subgoal images
via image editing as visual guidance. VPP~\cite{hu2025video} uses video diffusion
models to predict future visual states as intermediate representations for action
generation, and Seer~\cite{tian2025predictive} infers actions via inverse dynamics on
predicted latent visual features. Related WAMs explore further couplings between
visual prediction and action decoding~\cite{zhao2025cot,zhang2025upvla,zhu2025uwm,zhang2025dreamvla}.
WorldVLA~\cite{cen2025worldvla} formulates an autoregressive action-world model, and
Unified-VLA~\cite{wang2025unified} further unifies perception grounding,
vision-supervised world modeling, and action generation within a single framework.
Although these predictive intermediates make future visual-state structure explicit,
they live on the image plane rather than sharing the rigid-body geometry of the action
space, leaving the decoder to implicitly recover the rigid-body motion from these visual predictions.

\section{Problem Formulation and Analysis}
\label{sec:formulation}

\subsection{Problem Formulation}
\label{sec:task_definition}

At time step $t$, a visuomotor policy $\pi$ receives image observations $\mathbf{o}_t$,
a language instruction $l$, and the current end-effector state
$\mathbf{e}_t=[\mathbf{p}_t,\boldsymbol{\theta}_t]^\top$,
with position $\mathbf{p}_t\in\mathbb{R}^3$ and rotation parameterization
$\boldsymbol{\theta}_t$. Equivalently, $\mathbf{e}_t$ specifies a world-frame
homogeneous transformation,
\begin{equation}
\label{eq:state_matrix}
    \mathbf{T}_t =
        \begin{bmatrix}
            \mathbf{R}(\boldsymbol{\theta}_t) & \mathbf{p}_t \\
            \mathbf{0}^\top & 1
        \end{bmatrix} \in SE(3),
\end{equation}
where $\mathbf{R}(\boldsymbol{\theta}_t)\in SO(3)$ is the corresponding rotation matrix.

Let $\{\mathbf{T}_{t+h}\}_{h=1}^{H}$ denote a world-frame target pose sequence
over a prediction horizon of $H$ steps along a successful execution of the manipulation task. The policy must produce an
action chunk $\mathbf{A}_t=\{(\mathbf{a}_{t+h-1},g_{t+h-1})\}_{h=1}^{H}$ of the same length that
drives the end-effector through this pose sequence, where $\mathbf{a}_{t+h-1}$
is a 6-DoF relative end-effector action and $g_{t+h-1}$ is the gripper command.
For each horizon index $h$, the relative action is recovered in closed form as
\begin{equation}
\label{eq:action_recovery}
    \mathbf{a}_{t+h-1}
    = \rho\bigl(\mathbf{T}_{t+h-1}^{-1}\mathbf{T}_{t+h}\bigr),
\end{equation}

where $\rho:SE(3)\to\mathbb{R}^6$ parameterizes a relative transformation
as translation and rotation vectors, such as the axis-angle representation.
Equation~\eqref{eq:action_recovery} converts any target pose sequence into
executable 6-DoF relative actions. The gripper command is modeled separately
because opening and closing the gripper is not a rigid-body motion in $SE(3)$.

\subsection{Aligned Intermediate Representations of Visuomotor Policies}
\label{sec:geometric_inconsistency}

We model a visuomotor policy as a sequence of representation and
action-decoding stages and adopt the following design principle.
A horizon-indexed intermediate $\mathbf{m}_t=\{\mathbf{m}_{t+h}\}_{h=1}^{H}$
is \emph{geometrically aligned} with the action space if it provides
an explicit pose readout
$r(\mathbf{m}_{t+h})=\mathbf{g}\cdot\mathbf{T}_{t+h}\in SE(3)$ at each
horizon step, where $\mathbf{g}\in SE(3)$ is a fixed rigid transform that
may be unknown. Given such a readout, the decoder can approximate the
relative-action recovery of Eq.~\eqref{eq:action_recovery} while absorbing
$\mathbf{g}$, prediction noise, contact dynamics, and gripper timing as
residuals. We treat this as a design principle rather than a closed-form
sufficiency theorem, since successful execution still depends on the
residual control problem that the decoder must learn.

Vision-Language-Action (VLA) models, including feature-enhanced variants that inject
auxiliary spatial features, encode the observation, instruction, and end-effector
state into a latent multimodal feature and decode the action chunk directly,
\begin{equation}
\label{eq:vla_factorization}
    \mathbf{z}_t = E_{\mathrm{VLA}}(\mathbf{o}_t,l,\mathbf{e}_t),\qquad
    \mathbf{A}_t = D_{\mathrm{VLA}}(\mathbf{z}_t).
\end{equation}

Because $\mathbf{z}_t$ contains no explicit horizon-indexed pose readout,
$D_{\mathrm{VLA}}$ must implicitly infer both the target poses
$\{\mathbf{T}_{t+h}\}_{h=1}^{H}$ and the parameterization $\rho$ to realize
Eq.~\eqref{eq:action_recovery}. Pose recovery and action parameterization
are therefore coupled inside a single learned action decoder.

WAMs insert a predictive intermediate representation before action decoding,
\begin{equation}
\label{eq:wam_factorization}
    \mathbf{u}_t = F_{\mathrm{WAM}}(\mathbf{o}_t,l,\mathbf{e}_t)\in\mathcal{P}^{H},\qquad
    \mathbf{A}_t = D_{\mathrm{WAM}}(\mathbf{u}_t,\mathbf{e}_t),
\end{equation}

where $\mathcal{P}$ denotes an intermediate space such as future images
or latent visual features. Although $\mathbf{u}_t$ is horizon-indexed,
$\mathcal{P}$ does not share the rigid-body geometry of the action space.
The decoder $D_{\mathrm{WAM}}$ must therefore implicitly recover $\mathbf{T}_{t+h}$ from $\mathbf{u}_t$
before applying Eq.~\eqref{eq:action_recovery}. WAMs thus expose temporal
structure but leave the action-recovery step without geometric guidance.

OASIS instantiates this design principle by predicting a camera-frame
$SE(3)$ end-effector trajectory $\{\mathbf{T}^c_{t+h}\}_{h=1}^{H}$. This
predicted trajectory satisfies the geometric-alignment property with
$\mathbf{g}=\mathbf{T}_{c\to w}^{-1}$, the inverse of the camera-to-world
extrinsic, since
$\mathbf{T}^c_{t+h}=\mathbf{T}_{c\to w}^{-1}\cdot\mathbf{T}_{t+h}$.
The decoder therefore receives explicit pose information for action
generation up to this fixed rigid transform, and learns to absorb frame
conversion, prediction noise, contact dynamics, and gripper timing as
residuals.

\section{Methodology}
\label{sec:method}

\subsection{Overview of OASIS}
\label{sec:oasis_overview}

\begin{figure}
\centering
\includegraphics[width=\columnwidth]{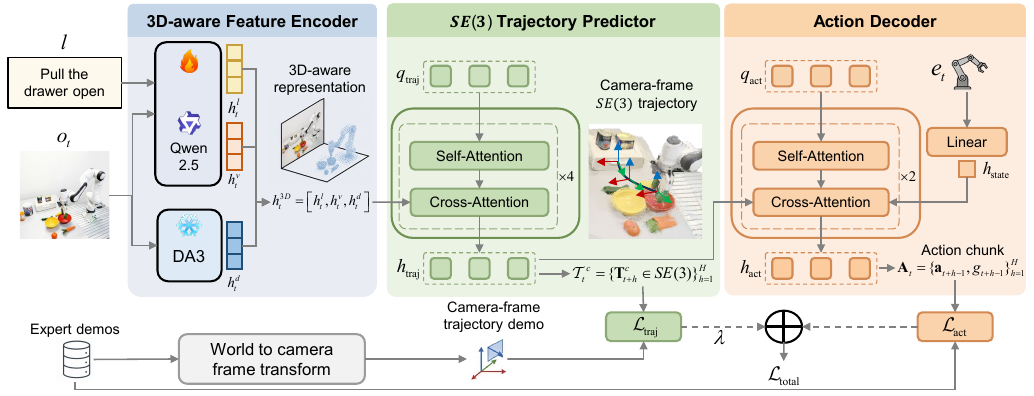}
\caption{Architecture of OASIS. The 3D-aware feature encoder merges image,
language, and metric-depth features into a 3D-aware representation. The
$SE(3)$ trajectory predictor aligns the intermediate representation with
the action space by predicting end-effector poses. The action decoder generates
executable action chunks conditioned on this aligned intermediate representation.}
\label{fig:method_overview}
\vspace{-8pt}
\end{figure}

OASIS instantiates the aligned-intermediate design principle from
Section~\ref{sec:geometric_inconsistency} with a three-stage policy architecture,
illustrated in Figure~\ref{fig:method_overview}.
Since robotic manipulation acts in 3D physical space while image
observations alone are 2D projections that lack metric depth, the 3D-aware feature encoder merges
vision-language and metric-depth features into a 3D-aware representation
$\mathbf{h}^{3D}_t$,
\begin{equation}
\label{eq:oasis_3d_rep}
    \mathbf{h}^{3D}_t = E_{3D}(\mathbf{o}_t,l).
\end{equation}
Conditioned on $\mathbf{h}^{3D}_t$, the trajectory predictor produces
pose-supervised hidden states $\mathbf{h}_{\mathrm{traj}}$, which a linear
head projects to a horizon-$H$ camera-frame end-effector $SE(3)$ trajectory
$\mathcal{T}^c_t$. This trajectory prediction aligns the predictor's
intermediate representation with the action space,
\begin{equation}
\label{eq:oasis_traj}
    \mathbf{h}_{\mathrm{traj}} = P_{SE(3)}(\mathbf{h}^{3D}_t), \qquad
    \mathcal{T}^c_t = \{\mathbf{T}^c_{t+h}\in SE(3)\}_{h=1}^{H} = \mathrm{Linear}(\mathbf{h}_{\mathrm{traj}}).
\end{equation}
Predicting in the camera frame keeps the intermediate in the same frame as
$\mathbf{h}^{3D}_t$ and spares the predictor from learning the
camera-to-world extrinsic. The action decoder, conditioned on these
pose-supervised hidden states $\mathbf{h}_{\mathrm{traj}}$ and the current
end-effector state $\mathbf{e}_t$, learns to approximate the relative-action
recovery of Eq.~\eqref{eq:action_recovery} and emit gripper commands,
\begin{equation}
\label{eq:oasis_decode}
    \mathbf{A}_t = D_{\mathrm{act}}(\mathbf{h}_{\mathrm{traj}},\mathbf{e}_t).
\end{equation}

\subsection{3D-aware Feature Encoder}
\label{sec:encoder}

The encoder provides the predictor with a 3D-aware representation that preserves
the language-conditioned visual information needed to forecast task-relevant target poses.

\textbf{Metric-depth feature.}
To supply the calibrated 3D structure that image observations lack, we use Depth
Anything~3~\cite{lin2025depth}, specifically the frozen DA3METRIC-LARGE model,
which extracts multi-scale metric-depth features from $\mathbf{o}_t$. We project
these features into $N_d$ depth hidden states $\mathbf{h}^d_t\in\mathbb{R}^{N_d\times D}$, where $D$ is the hidden dimension shared by the encoder, predictor, and decoder.

\textbf{Vision-language feature.}
We use a VLM with a Qwen2.5-0.5B
backbone~\cite{bai2025qwen2,karamcheti2024prismatic} that jointly processes
$\mathbf{o}_t$ and $l$, producing $N_v$ visual hidden states
$\mathbf{h}^v_t\in\mathbb{R}^{N_v\times D}$ and $N_l$ language hidden states
$\mathbf{h}^l_t\in\mathbb{R}^{N_l\times D}$. We concatenate $\mathbf{h}^l_t$,
$\mathbf{h}^v_t$, and $\mathbf{h}^d_t$ into the 3D-aware representation of Eq.~\eqref{eq:oasis_3d_rep},
\begin{equation}
\label{eq:encoder_concat}
    \mathbf{h}^{3D}_t = [\mathbf{h}^l_t,\mathbf{h}^v_t,\mathbf{h}^d_t].
\end{equation}

\subsection{$SE(3)$ Trajectory Predictor}
\label{sec:trajectory}

The trajectory predictor aligns its intermediate representation with the
action space by predicting the trajectory $\mathcal{T}^c_t$ of Eq.~\eqref{eq:oasis_traj},
where each $\mathbf{T}^c_{t+h}\in SE(3)$ is the predicted end-effector pose
at horizon step $h$. A linear head projects the predictor's hidden states
$\mathbf{h}_{\mathrm{traj}}$ to this trajectory. These pose-supervised
hidden states serve as the conditioning signal passed to the action decoder.

\textbf{Architecture.}
The trajectory predictor initializes $H$ learnable trajectory queries
$\mathbf{q}_{\mathrm{traj}}\in\mathbb{R}^{H\times D}$. Stacked transformer
blocks process these queries with self-attention and Rotary Position Embeddings
(RoPE)~\cite{SU2024127063} for temporal coherence, while cross-attention
conditions each query on $\mathbf{h}^{3D}_t$. The resulting trajectory hidden
states $\mathbf{h}_{\mathrm{traj}}\in\mathbb{R}^{H\times D}$ are projected to
per-step pose vectors
\begin{equation}
\label{eq:traj_vectors}
    \boldsymbol{\tau}^c_t =
    \{\mathbf{e}^c_{t+h}=[\mathbf{p}^c_{t+h},\boldsymbol{\theta}^c_{t+h}]^\top
    \}_{h=1}^{H},
\end{equation}
where $\mathbf{p}^c_{t+h}\in\mathbb{R}^3$ is the position and
$\boldsymbol{\theta}^c_{t+h}\in\mathbb{R}^3$ is an axis-angle vector.
Axis-angle is the canonical exponential-coordinate chart of $SO(3)$. Every
prediction $\boldsymbol{\theta}^c_{t+h}$ yields a valid rotation matrix
$\mathbf{R}(\boldsymbol{\theta}^c_{t+h})\!=\!\exp([\boldsymbol{\theta}^c_{t+h}]_\times)\!\in\!SO(3)$
in closed form via the Rodrigues rotation formula, where $[\cdot]_\times$ maps a vector to its skew-symmetric matrix, so each pose vector induces a valid
$SE(3)$ element
\begin{equation}
    \label{eq:camera_state_matrix}
    \mathbf{T}^c_{t+h} =
    \begin{bmatrix}
        \mathbf{R}(\boldsymbol{\theta}^c_{t+h}) & \mathbf{p}^c_{t+h} \\
        \mathbf{0}^\top & 1
    \end{bmatrix}\in SE(3)
\end{equation}
by construction, requiring no projection, Gram-Schmidt
orthogonalization, or post-hoc correction. For the bounded range of
camera-frame end-effector orientations encountered in tabletop
manipulation we further have $\|\boldsymbol{\theta}^c_{t+h}\|<\pi$,
keeping all predictions inside the canonical axis-angle chart of $SO(3)$
and the network output on the $SE(3)$ manifold throughout training and
inference.

\textbf{Training objective.}
For computational efficiency, we supervise the local pose parameterization with
ground-truth pose vectors
$\hat{\boldsymbol{\tau}}^c_t=\{\hat{\mathbf{e}}^c_{t+h}\}_{h=1}^{H}$ obtained by
transforming world-frame demonstrations into the camera frame, and adopt the
trajectory loss
\begin{equation}
    \label{eq:traj_loss}
    \mathcal{L}_{\mathrm{traj}}
    = \frac{1}{H}\sum_{h=1}^{H}
    \left\|\hat{\mathbf{e}}^c_{t+h}-\mathbf{e}^c_{t+h}\right\|_1.
\end{equation}

We choose $\ell_1$ because it is robust to teleoperation artifacts and
yields precise pose modeling~\cite{zhao2023learning}. Empirically, axis-angle outperforms the quaternion and Euler parameterizations under matched training on LIBERO-Long, as Figure~\ref{fig:rotation_param} shows. In our setting, target and predicted absolute future orientations remain within the canonical axis-angle chart, away from the $\|\boldsymbol{\theta}^c_{t+h}\| = \pi$ singular set, so axis-angle coordinates are well-defined throughout training and inference.

\subsection{Action Decoder}
\label{sec:action}

The action decoder cross-attends to the predictor's pose-supervised hidden
states and the current end-effector state, approximating the relative-action
recovery of Eq.~\eqref{eq:action_recovery} while absorbing frame conversion,
prediction noise, contact dynamics, and gripper timing as residuals. We
retain a learned decoder rather than a closed-form pose-to-action converter
because these residuals exceed what explicit pose readouts can capture, as
detailed in Appendix~\ref{appendix_c3:closed_form_decoder}.

\textbf{Architecture.}
The action decoder initializes $H$ learnable action queries
$\mathbf{q}_{\mathrm{act}}\in\mathbb{R}^{H\times D}$. These queries attend
to two sources of context, namely the trajectory hidden states
$\mathbf{h}_{\mathrm{traj}}$ from the predictor and a state embedding
$\mathbf{h}_{\mathrm{state}}\in\mathbb{R}^{1\times D}$ obtained by
linearly projecting the current end-effector state $\mathbf{e}_t$. The
two are concatenated into context hidden states
$\mathbf{h}_{\mathrm{ctx}} = [\mathbf{h}_{\mathrm{traj}}, \mathbf{h}_{\mathrm{state}}] \in \mathbb{R}^{(H+1) \times D}$.
The decoder adds no encoding of the order in which it receives the $H$ trajectory hidden states, so its cross-attention over them is permutation-invariant.
The decoder then outputs the action chunk
$\mathbf{A}_t=\{(\mathbf{a}_{t+h-1},g_{t+h-1})\}_{h=1}^{H}$, where
$\mathbf{a}_{t+h-1}=[\Delta\mathbf{p}_{t+h-1},\Delta\boldsymbol{\theta}_{t+h-1}]^\top$
is the 6-DoF relative action and $g_{t+h-1}$ is the gripper command.

\textbf{Training objective.}
The action loss supervises the output action chunk against expert demonstrations,
\begin{equation}
    \label{eq:action_loss}
    \mathcal{L}_{\mathrm{act}}
    = \frac{1}{H}\sum_{h=1}^{H}
    \left\|
    [\hat{\mathbf{a}}_{t+h-1},\hat{g}_{t+h-1}]
    - [\mathbf{a}_{t+h-1},g_{t+h-1}]
    \right\|_1.
\end{equation}
The total objective combines trajectory supervision with executable-action supervision,
\begin{equation}
    \label{eq:total_loss}
    \mathcal{L}_{\mathrm{total}}
    = \lambda\mathcal{L}_{\mathrm{traj}}+\mathcal{L}_{\mathrm{act}},
\end{equation}
where $\lambda$ balances the two losses and is set to $0.1$.

\section{Experiments}
\label{sec:experiments}

We evaluate OASIS in simulation and on real robots to answer four questions.
First, does OASIS surpass VLA and WAM baselines on standard manipulation suites? Second, is the
metric-depth feature of the 3D-aware feature encoder necessary? Third, how do the supervision target of the predictor's hidden states and its
reference frame affect policy success? Fourth, does OASIS's lead transfer
to real robots across multi-task suites and under OOD
perturbations?

\subsection{Implementation Details}
\label{sec:implementation_details}
The trajectory predictor and action decoder share a transformer architecture
with four and two blocks respectively. The predictor outputs an eight-step
$SE(3)$ end-effector trajectory in the camera frame, and the decoder
generates the corresponding action chunk conditioned on the trajectory
hidden states. OASIS has 1.73B total parameters, of which 0.18B are
trainable. We LoRA-tune~\cite{hu2022lora} the Qwen2.5-0.5B VLM, train the
predictor, decoder, and projection layers from scratch without large-scale
robotic pretraining, and freeze DA3METRIC-LARGE~\cite{lin2025depth}. Experiments use four
NVIDIA A800 GPUs with batch size 64.
Appendix~\ref{appendix_b:implementation_details} provides further details.

\subsection{Simulation Experiments}
\label{sec:simulation_experiments}

\textbf{Simulation setup.} To answer the first question, we evaluate OASIS on
two widely adopted simulation benchmarks, LIBERO~\cite{liu2023libero} and
CALVIN~\cite{mees2022calvin}. LIBERO has four task suites, namely Spatial,
Object, Goal, and Long, which test spatial reasoning, object grounding, goal-directed behavior, and
long-horizon planning. Each contains 10 tasks with 50 human-teleoperated demonstrations per task,
and we evaluate every task over 50 episodes. For CALVIN, we use the challenging ABC$\to$D setting to test
OOD generalization to an unseen environment D after training on environments
A, B, and C, reporting the average sequence length over 1{,}000 distinct
instruction chains, each of five consecutive subtasks. Detailed descriptions of the simulation environments
appear in Appendix~\ref{appendix_c1:simulation_settings}.

\textbf{Evaluation protocol.} The evaluation protocol for LIBERO and CALVIN
uses only RGB images, language, and robot state as input. We follow this protocol,
train OASIS for 50k steps per suite, and average across three seeds. Baseline numbers
come from the original papers. Their tasks and benchmark splits match ours, while evaluation counts, optimization, and model selection follow each source.
We primarily compare against RGB-only baselines, and include 3D Diffuser Actor~\cite{ke20243d}
as a structured-pose policy in Table~\ref{table:calvin_benchmark},
although it consumes calibrated multi-view RGB-D.

\begin{table}[t]
  \caption{Success rates in percent on the four LIBERO suites. The Pretrain
  column marks pre-training on large-scale robotic
  datasets, and the Intermediate column lists the representation each method
  feeds into the action decoder. The highest success rate in each suite is
  in bold.
  OASIS reaches the highest average success at 97.6 without any pre-training on large-scale robotic datasets.}
  \label{table:libero_benchmark}
  {\centering
  \small
  \setlength{\tabcolsep}{5pt}
  \begin{tabular}{lccccccc}
    \toprule
    \textbf{Method} & \textbf{Intermediate} & \textbf{Pretrain} & \textbf{Spatial} & \textbf{Object} &
    \textbf{Goal} & \textbf{Long} & \textbf{Average} \\
    \midrule
    SpatialVLA~\cite{qu2025spatialvla}        & spatial features & \yesmark & 88.2          & 89.9          & 78.6          & 55.5          & 78.1 \\
    WorldVLA~\cite{cen2025worldvla}           & future visual states     & \nomark & 85.6          & 89.0          & 82.6          & 59.0          & 79.1 \\
    ThinkAct~\cite{huang2025thinkact}         & 2D-supervised features & \yesmark & 88.3          & 91.4          & 87.1          & 70.9          & 84.4 \\
    $\pi_0$~\cite{BlackK-RSS-25}              & multimodal features & \yesmark & 96.8          & \textbf{98.8} & 95.8          & 85.2          & 94.2 \\
    QDepth-VLA~\cite{li2025qdepth}            & spatial features & \yesmark & 97.6          & 96.6          & 95.2          & 90.0          & 94.9 \\
    UniVLA~\cite{BuQ-RSS-25}                  & spatial features & \yesmark & 96.5          & 96.8          & 95.6          & 92.0          & 95.2 \\
    Unified-VLA~\cite{wang2025unified}        & future visual states & \yesmark & 95.4          & \textbf{98.8} & 93.6          & 94.0          & 95.5 \\
    \oursmidrule
    \textbf{OASIS (Ours)} & SE(3)-supervised features & \nomark & \textbf{99.0} & \textbf{98.8} & \textbf{97.4} & \textbf{95.2} & \textbf{97.6} \\
    \oursbottomrule
  \end{tabular}\par}
  \caption{Results on CALVIN ABC$\to$D.
  The Pretrain column marks pre-training on
  large-scale robotic datasets. Tasks completed in a row measures the success rate in percent of
  sequentially executing 1 to 5 instructions without scene resets. Avg. is the average sequence
  length, the mean number of instructions completed in a row. $^{\dagger}$ uses multi-view RGB-D.
  OASIS attains the highest average sequence length at 4.57, with its largest gains at the fourth and fifth instructions.}
  \label{table:calvin_benchmark}
  \centering
  \small
  \setlength{\tabcolsep}{4.5pt}
  \begin{tabular}{lcccccccc}
    \toprule
    \textbf{Method} & \textbf{Intermediate} & \textbf{Pretrain} &
    \multicolumn{5}{c}{\textbf{Tasks Completed in a Row}} & \textbf{Avg.} \\
    \cmidrule(lr){4-8}
     & & & 1 & 2 & 3 & 4 & 5 & \\
    \midrule
    SuSIE~\cite{black2023zero}             & future visual states & \nomark & 87.0         & 69.0          & 49.0          & 38.0          & 26.0          & 2.69 \\
    3D Diffuser Actor$^{\dagger}$~\cite{ke20243d} & 3D feature & \nomark & 93.8          & 80.3          & 66.2          & 53.3          & 41.2          & 3.35 \\
    ReconVLA~\cite{song2026reconvla}       & spatial features & \yesmark & 95.6          & 87.6          & 76.9          & 69.3          & 64.1          & 3.95 \\
    Seer-Large~\cite{tian2025predictive}   & future visual states & \yesmark & 96.3          & 91.6          & 86.1          & 80.3          & 74.0          & 4.28 \\
    VPP~\cite{hu2025video}                 & future visual states & \yesmark & 96.5          & 90.9          & 86.6          & 82.0          & 76.9          & 4.33 \\
    Unified-VLA~\cite{wang2025unified}     & future visual states & \yesmark & \textbf{98.9} & 94.8          & 89.0          & 82.8          & 75.1          & 4.41 \\
    DreamVLA~\cite{zhang2025dreamvla}      & future visual states & \yesmark & 98.2          & 94.6          & 89.5          & 83.4          & 78.1          & 4.44 \\
    \oursmidrule
    \textbf{OASIS (Ours)} & SE(3)-supervised features & \nomark & 98.1 & \textbf{94.9} & \textbf{91.7} & \textbf{88.9} & \textbf{83.3} & \textbf{4.57} \\
    \oursbottomrule
  \end{tabular}
\end{table}

\textbf{LIBERO simulation results.} As reported in Table~\ref{table:libero_benchmark},
OASIS reaches a 97.6\% average success rate on LIBERO, leading the next-best baseline
Unified-VLA~\cite{wang2025unified} by 2.1 points. OASIS leads or ties on all four suites, with margins of
1.4 points on Spatial, 1.6 on Goal, and 1.2 on Long over the strongest prior result on
each suite, and a tie on Object. Appendix~\ref{appendix_c5:seed_variance} lists the standard deviations, and even the worst Long seed reaches 94.8\%, above Unified-VLA at 94.0\%.
Against the strongest model of each family, OASIS leads the vanilla VLA $\pi_0$ by 3.4 points, the
feature-enhanced UniVLA by 2.4, and the WAM Unified-VLA by 2.1. The only other method without
large-scale robotic pretraining, WorldVLA, reaches 79.1\%.
ThinkAct~\cite{huang2025thinkact}, evaluated under the same protocol, predicts a 2D image-plane
trajectory, and OASIS leads it by 13.2 points on average and 24.3 on Long.

\textbf{CALVIN simulation results.} As reported in Table~\ref{table:calvin_benchmark},
OASIS attains an average sequence length of 4.57 on CALVIN ABC$\to$D, ahead of every
baseline. It trails Unified-VLA and DreamVLA on the first instruction and leads from the
second onward, and it exceeds the multi-view RGB-D policy 3D Diffuser Actor by 1.22 in average
sequence length. The gap widens on the five-consecutive-task setting, where OASIS reaches
83.3\% and leads the next-best baseline DreamVLA~\cite{zhang2025dreamvla} by 5.2 points. This widening lead suggests that small per-step errors accumulate less
under an intermediate aligned via $SE(3)$ trajectory prediction than
under intermediates that live in the observation space.

\begin{figure}[t]
  \centering
  \includegraphics[width=\linewidth]{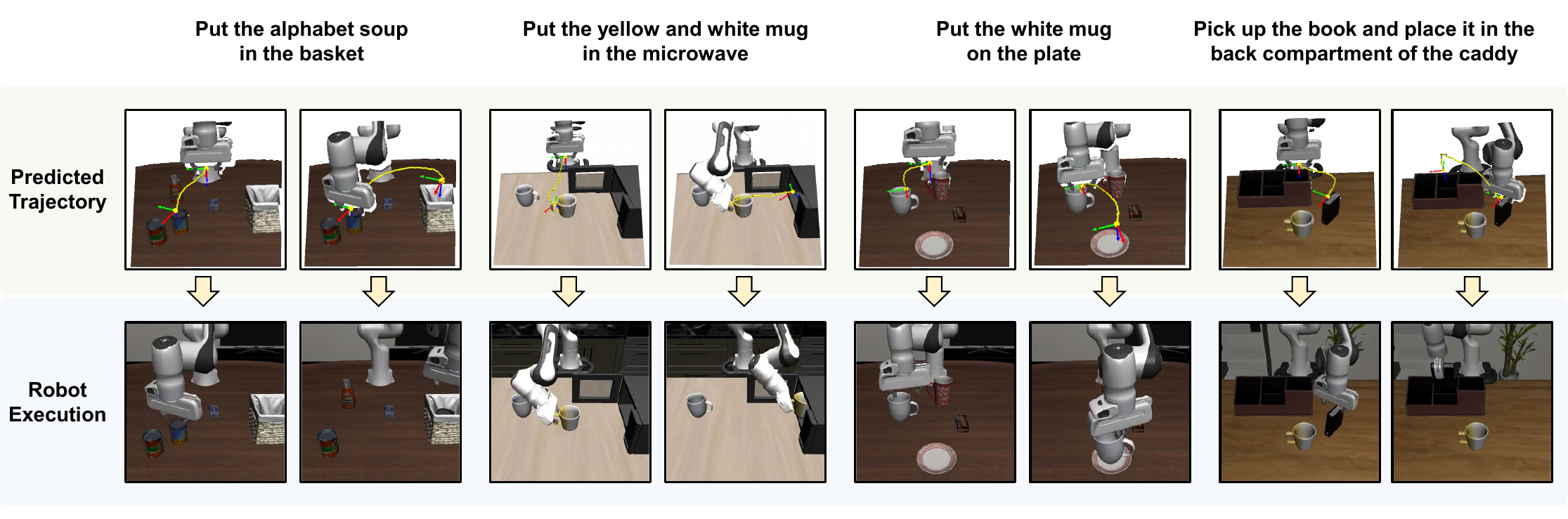}
  \captionof{figure}{Predicted $SE(3)$ trajectories and robot execution on four LIBERO tasks. The top row
  overlays the predicted waypoints and rotation axes, and the bottom row shows the executed motion.}
  \label{fig:visualization}
  \vspace{4pt}
  \includegraphics[width=\linewidth]{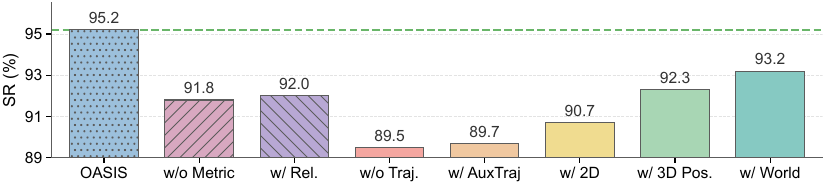}
  \captionof{figure}{Ablation summary on LIBERO-Long. Success increases across the tested supervision ladder, and the trajectory predictor has the largest effect on success, as removing it costs 5.7 points.}
  \label{fig:ablation_bar_long}
  \vspace{4pt}
  \begin{minipage}[t]{0.5182\textwidth}
    \centering
    \includegraphics[width=\linewidth]{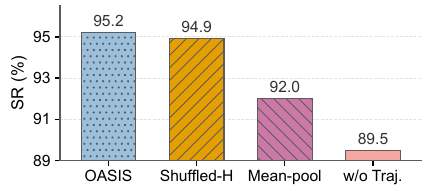}
    \caption{Decoder conditioning controls, averaged over three seeds on LIBERO-Long.}
    \label{fig:decoder_controls}
  \end{minipage}\hfill
  \begin{minipage}[t]{0.4182\textwidth}
    \centering
    \includegraphics[width=\linewidth]{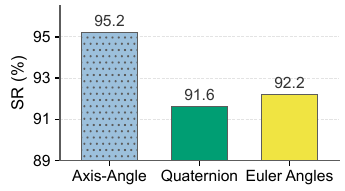}
    \caption{Ablation on choices of $SO(3)$ parameterization on LIBERO-Long.}
    \label{fig:rotation_param}
  \end{minipage}
\end{figure}

\textbf{Visualization of the $SE(3)$ trajectory.} To validate the trajectory predictor
qualitatively, we plot in Figure~\ref{fig:visualization} the predicted translation
waypoints and rotation axes alongside the executed end-effector path on four complex
tasks. The executed path closely follows the predicted positions and orientations,
indicating that the predicted trajectory is a consistent rigid-body prior for action generation.

\subsection{Ablation Study}
\label{sec:ablation_study}

To answer questions two and three, we ablate the metric depth, the $SE(3)$ trajectory
predictor, and the $SO(3)$ parameterization on LIBERO-Long. The first two are
repeated on LIBERO-Spatial in Appendix~\ref{appendix_c2:more_ablation}, where the
ranking among variants is consistent.

\textbf{Effectiveness of the 3D-aware feature encoder.} The two hatched
bars adjacent to the OASIS bar in Figure~\ref{fig:ablation_bar_long}
ablate the metric-depth feature. The variant \textit{w/o Metric}, which removes this
feature from the 3D-aware representation, drops
LIBERO-Long from 95.2\% to 91.8\%, and the variant \textit{w/ Rel.},
which substitutes the relative depth of Depth Anything V2, recovers
only marginally to 92.0\%. Metric scale, not depth in general, is what
helps the 3D-aware representation.

\textbf{Impact of $SE(3)$ trajectory prediction.}
Figure~\ref{fig:ablation_bar_long} sweeps the supervision target on the
pose-supervised hidden states under a matched-backbone control, with all
variants sharing the encoder, depth model, and training
budget. Richer geometric targets yield a monotonic increase on
LIBERO-Long from 89.5\% to 95.2\%. The variant \textit{w/o Traj.} drops
the trajectory predictor and reaches 89.5\%, marking the floor of the
ladder. Since it keeps the same backbone and frozen depth model, \textit{w/o Traj.} is itself a depth-enhanced VLA, and removing the trajectory predictor costs most among the ablated components, 5.7 points against 3.4 for removing metric depth.
\textit{w/ AuxTraj} retains the trajectory loss but routes it
through a parallel predictor branch, so only $\mathbf{h}^{3D}_t$, not
$\mathbf{h}_{\mathrm{traj}}$, conditions the decoder. It climbs only to
89.7\%, showing that pose supervision helps only when the supervised
hidden states reach the decoder. \textit{w/ 2D} reaches 90.7\%, barely
above \textit{w/o Traj.}, since image-plane geometry lacks the rigid-body
structure of the action space. Adding depth lifts \textit{w/ 3D Pos.} to
92.3\%, and adding rotation lifts \textit{w/ World} to 93.2\%, 2.0 points
below the camera-frame counterpart. The camera-frame $SE(3)$ trajectory attains the
highest success, consistent with camera-frame prediction sparing the predictor
from recovering the camera-to-world extrinsic.

\textbf{How the decoder uses the trajectory hidden states.}
Figure~\ref{fig:decoder_controls} alters the trajectory hidden states of the trained OASIS model at evaluation time and compares it with \textit{w/o Traj.}
Since the decoder adds no encoding of their order, permuting them in \textit{Shuffled-H} changes the actions only by summation-order noise, at most $7.8\times10^{-3}$, and success nearly matches OASIS.
\textit{Mean-pool} replaces the $H$ states by their mean, keeping the trajectory-level mean but removing step-wise variation.
Its 3.2-point drop shows that the decoder uses step-wise content, and its 2.5-point lead over \textit{w/o Traj.} is consistent with useful trajectory-level conditioning.
As Appendix~\ref{appendix_c4:decoder_attention} and Table~\ref{tab:decoder_attention} detail, one attention head of the first decoder block puts 0.98 to 1.00 of each action step's weight on a single trajectory hidden state, and the last two steps read each other's states, so this one-to-one mapping follows content rather than position.
From \textit{w/ 2D} to OASIS, every variant carries a trajectory loss while success rises with the geometric richness of the target, so it is the step-wise content, not the trajectory loss itself, that drives success.
Appendix~\ref{appendix_c3:closed_form_decoder} further shows that replacing the learned decoder with a hardcoded pipeline, which converts the predicted poses into actions, collapses LIBERO-Long to 0.0\%, confirming the decoder is necessary to absorb prediction noise, contact dynamics, and gripper timing.

\textbf{Choice of $SO(3)$ parameterization.} Figure~\ref{fig:rotation_param} compares
three $SO(3)$ parameterizations under matched training. Axis-angle reaches 95.2\%, 3.0 points above
Euler angles and 3.6 above quaternions. This ranking is consistent with axis-angle being
a minimal, constraint-free chart that covers the bounded range of camera-frame
end-effector orientations encountered in tabletop manipulation.

\begin{figure}[t]
  \centering
  \includegraphics[width=\linewidth]{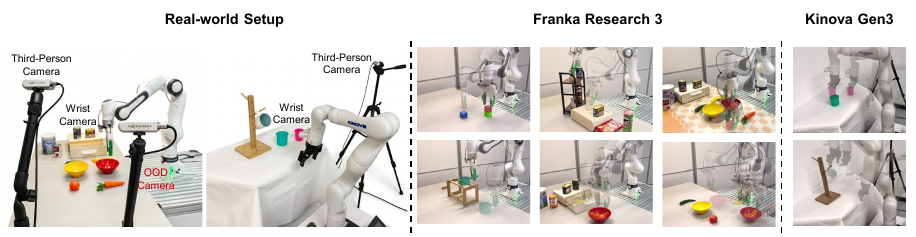}
  \captionof{figure}{Real-world platforms and task examples. Each robot uses a third-person and a wrist
  camera, and OOD Camera marks where the third-person camera is moved for the displaced-camera test.}
  \label{fig:real_world_platform}

  \begin{minipage}[c]{0.50\textwidth}
    \centering
    \small
    \setlength{\tabcolsep}{4pt}
    \captionof{table}{Real-world success rates in percent on the Goal, Spatial, and Long suites,
    with each task evaluated over three independent runs of 20 trials.}
    \label{table:multitask_results}
    \begin{tabular}{lcccc}
      \toprule
      \textbf{Method} & \textbf{Goal} & \textbf{Spatial} & \textbf{Long} & \textbf{Average} \\
      \midrule
      ACT~\cite{zhao2023learning}                   & 58.3          & 45.0          & 18.3            & 40.5 \\
      Seer-Large~\cite{tian2025predictive}            & 73.3          & 55.2          & 46.7            & 58.4 \\
      RDT~\cite{liu2025rdtb}                   & 81.7          & 66.7          & 60.0          & 69.5 \\
      $\pi_{0.5}$~\cite{pmlr-v305-black25a}          & 95.0          & 78.3          & 71.6            & 81.6 \\
      \oursmidrule
      \textbf{OASIS (Ours)} & \textbf{98.6}  & \textbf{85.8} & \textbf{83.3}   & \textbf{89.2}
      \\
      \oursbottomrule
    \end{tabular}
  \end{minipage}\hfill
  \begin{minipage}[c]{0.45\textwidth}
    \centering
    \includegraphics[width=\linewidth]{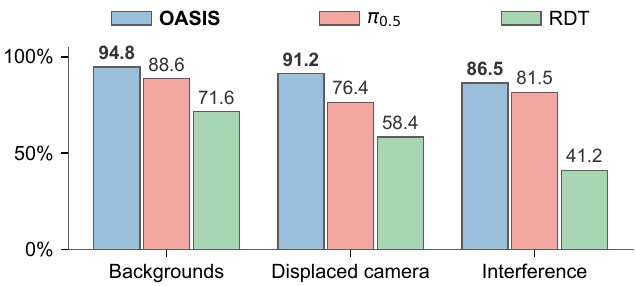}
    \captionof{figure}{Success rates on the Goal suite in three OOD
    settings. OASIS loses less than $\pi_{0.5}$ under each of the three perturbations.}
    \label{fig:ood_results}
  \end{minipage}
\end{figure}

\subsection{Real-world Experiments}
\label{sec:real_world_experiments}

\textbf{Real-world setup.} To answer the fourth question, we evaluate OASIS on a
Franka Research 3 and a Kinova Gen3, each with a third-person and a wrist camera, shown in Figure~\ref{fig:real_world_platform}.
Three suites probe instruction following, spatial reasoning, and long-horizon manipulation,
denoted Goal, Spatial, and Long. OASIS and the compared baselines are trained on
50 teleoperated demonstrations per task and evaluated over three independent runs of
20 trials each, for 60 trials per task, exceeding the trial counts reported in the
original baseline papers. All methods are trained for 50k steps and evaluated at the final checkpoint without hyperparameter search, and only $\pi_{0.5}$, RDT, and Seer-Large start from robotic-pretrained weights.
OOD perturbations are further tested on the Goal suite
under unseen backgrounds, a third-person camera displaced by about 15\,cm, and human interference. Full
settings and task descriptions appear in Appendix~\ref{appendix_d1:real_world_experiments_settings}.

\textbf{Main results analysis.} As reported in Table~\ref{table:multitask_results}, OASIS
reaches an 89.2\% average success rate, including 98.6\% on Goal, 85.8\% on Spatial, and
83.3\% on Long, leading $\pi_{0.5}$~\cite{pmlr-v305-black25a} by 7.6 points, RDT~\cite{liu2025rdtb} by 19.7, Seer-Large~\cite{tian2025predictive} by 30.8, and ACT~\cite{zhao2023learning}
by 48.7, with the numerical lead holding across all three suites, although Appendix~\ref{appendix_d3:wilson_cis} shows that the 95\% Wilson intervals against $\pi_{0.5}$ overlap, widely on Long and slightly on Goal and Spatial.
The numerical margin over $\pi_{0.5}$ is largest on Long, 11.7 points against 3.6 on Goal and 7.5 on
Spatial.

\textbf{OOD results analysis.} As shown in Figure~\ref{fig:ood_results}, OASIS holds
94.8\%, 91.2\%, and 86.5\% across unseen backgrounds, displaced camera, and human
interference, while $\pi_{0.5}$ reaches 88.6\%, 76.4\%, and 81.5\%. Under unseen
backgrounds, the metric-depth feature keys off scene geometry, so distractor objects and tabletops barely shift the decoder input.
Under the displaced camera, only the third-person view moves, and because
$\mathbf{h}^{3D}_t$ and $\mathcal{T}^c_t$ both live in the camera frame, the
decoder tolerates this displacement somewhat.
This probes robustness to a single uncalibrated viewpoint shift and does not establish camera-pose invariance.
Under human
interference, the predictor regenerates the trajectory for each chunk, so OASIS
retargets within one action chunk when the bowl is moved mid-execution.
The residual robustness of $\pi_{0.5}$ and RDT likely reflects their
large-scale robotic pretraining.

\textbf{Failures, speed, and data efficiency.} A failure analysis in Appendix~\ref{appendix_d2:real_world_experiments_results}
traces OASIS's residual errors
to sub-centimeter placement and long-horizon compounding, and an error decomposition of five tasks shows
that on each platform the lowest-success task among them has the most accurate predicted trajectory,
pointing to execution tolerance rather than prediction error. One inference takes 45\,ms per
action chunk on one RTX 4090, against 120\,ms for $\pi_{0.5}$ and 260\,ms for RDT.
A data-scaling study in the same appendix
shows that on the real-world Long task, OASIS reaches 35.0\% with only 10 demonstrations, the success
rate that $\pi_{0.5}$ reaches with 25, a low-data advantage on
this task.

\section{Conclusion}
\label{sec:conclusion}

Existing visuomotor policies decode 6-DoF actions from intermediate
representations that do not share the rigid-body geometry of the action
space. We address this gap by aligning the intermediate representation
with the action space via camera-frame $SE(3)$ end-effector trajectory
prediction, instantiated as OASIS, whose action decoder cross-attends to
the predictor's pose-supervised hidden states. Across simulation benchmarks and real-world
platforms, OASIS attains superior manipulation precision and generalization,
suggesting that aligning the intermediate is a more direct lever for
visuomotor policy performance than enriching observation-space features.

\textbf{Limitations and future work.}\label{sec:limitations}
Our study is limited to tabletop manipulation with a single fixed-base arm,
and OASIS has not been validated on bimanual or mobile manipulation tasks.
Since the $SE(3)$ intermediate aligns with the action space of a single
end-effector, extending this design principle to richer action spaces is a
natural next step, including coupled pose manifolds for bimanual
manipulation, coupled $SE(2) \times SE(3)$ trajectories for mobile
manipulation, and contact-conditioned trajectories for dexterous hands.
Future work will also examine the choice between regression and generative
trajectory prediction, such as diffusion or flow matching, which the
alignment principle equally admits since it concerns what the intermediate
represents rather than how it is modeled.

\begin{ack}
This work was supported in part by the NSFC under Grant No.~62125305, the Fundamental and Interdisciplinary Disciplines Breakthrough Plan of the Ministry of Education of China under Grant No.~JYB2025XDXM210, the NSFC under Grant Nos.~U23A20339, 62573339, and 62503380, the Natural Science Foundation of Shaanxi Province under Grant No.~\mbox{2025SYSSYSZD-083}, the State Grid Corporation of China Science and Technology Projects \mbox{52060025005B-439-ZN}, and the Shaanxi Provincial Key R\&D Program under Grant No.~\mbox{2026JC-ZDGG-17}.
\end{ack}

\bibliographystyle{plainnat}
\bibliography{reference}

\newpage
\appendix
\setcounter{topnumber}{3}
\setcounter{totalnumber}{4}
\section{Implementation details}
\label{appendix_b:implementation_details}
\subsection{OASIS architecture}
\label{appendix_b1:architecture}

Given the real-time constraints of visuomotor policies, we
prioritize a lightweight pretrained VLM over heavily parameterized alternatives.
As Table~\ref{table:param_breakdown} shows, the trajectory predictor and action decoder together comprise only 70 million
parameters, keeping training and inference efficient.

\paragraph{Vision-language model.} OASIS uses a VLM architecture derived from Prismatic
VLM~\cite{karamcheti2024prismatic}, with Qwen2.5-0.5B~\cite{bai2025qwen2} as the language backbone and two visual
encoders, DINOv2~\cite{oquab2024dinov} and SigLIP~\cite{zhai2023sigmoid}. The VLM is
pretrained on the LLaVA-1.5-Instruct dataset~\cite{liu2023visual} and supplies the visual and
language hidden states of the 3D-aware representation in Section~\ref{sec:encoder}.

\paragraph{Metric-depth module.} OASIS uses the DA3METRIC-LARGE model of Depth Anything 3~\cite{lin2025depth}
to supply metric depth. Unlike Depth Anything
V2~\cite{yang2024depth} and VGGT~\cite{wang2025vggt}, which produce normalized relative
depth maps, this model estimates depth in physical units, the absolute scale that precise
manipulation needs and that the metric-depth ablation in Section~\ref{sec:ablation_study} isolates.

\paragraph{Trajectory predictor.} The trajectory predictor is a 4-block
Transformer with $H=8$ learnable trajectory query tokens. Within
each block, these queries perform self-attention with Rotary Position Embeddings, followed
by cross-attention whose keys and values come from the 3D-aware
representation $\mathbf{h}^{3D}_t$ of language, visual, and metric-depth hidden
states. The predictor outputs the trajectory hidden states $\mathbf{h}_{\mathrm{traj}}$, from which a
linear head decodes the predicted $SE(3)$ trajectory. The hidden states condition the action
decoder, and the decoded trajectory receives the loss of Section~\ref{sec:trajectory}.

\paragraph{Action decoder.} The action decoder adopts the same architecture as the
trajectory predictor but with 2 blocks. Its cross-attention takes the trajectory hidden states and the
embedding of the current end-effector state as keys and values, so the decoder generates actions
conditioned on the trajectory hidden states rather than on the decoded trajectory, as
Appendix~\ref{appendix_c4:decoder_attention} examines. Table~\ref{parameter} summarizes the
configurations of the trajectory predictor and action decoder.

\begin{table}[!htbp]
  \caption{Full parameter breakdown of OASIS.}
  \label{table:param_breakdown}
  \centering
  \begin{tabular}{lcc}
    \toprule
    \textbf{Component} & \textbf{Total (B)} & \textbf{Trainable (B)} \\
    \midrule
    DA3METRIC-LARGE                               & 0.35          & 0.00 (Frozen) \\
    Qwen2.5-0.5B VLM                              & 1.30          & 0.10 (LoRA) \\
    Trajectory predictor + action decoder         & 0.07          & 0.07 \\
    Linear projections + state embedding          & 0.01          & 0.01 \\
    \midrule
    \textbf{Total}                                & \textbf{1.73} & \textbf{0.18} \\
    \bottomrule
  \end{tabular}
\end{table}

\begin{table}[!htbp]
	\centering
	\caption{Architecture and parameters of the trajectory predictor and action decoder.}
    \label{parameter}%
	\begin{tabular}{ccc}
		\toprule
		\textbf{Module} & \textbf{Trajectory Predictor} & \textbf{Action Decoder}\\
        \midrule
        Layers  & 4 & 2 \\
		Hidden Size  & 896 & 896   \\
		Attention Heads & 8 & 8      \\
		Number of Queries ($H$)  & 8 & 8\\
		\midrule
      Total Parameters & 0.05B & 0.02B \\
		\bottomrule
	\end{tabular}%
\end{table}%

\subsection{Training details}
\label{appendix_b2:training_detail}

\paragraph{Training dataset.} We use datasets in the standard RLDS format. Inputs consist of a
language instruction, RGB images from the third-person and wrist views resized to $224 \times
224 \times 3$, and a 7-dimensional robot state comprising the end-effector position,
rotation, and gripper state. The policy is supervised with 6-DoF relative actions and gripper
commands. Trajectory labels are world-frame end-effector poses transformed into the third-person
camera frame with its extrinsic matrix.

\paragraph{Training settings.} We train OASIS with AdamW at a learning rate of $2{\times}10^{-4}$
under a cosine annealing schedule with 5,000 warm-up steps. The vision-language model is
fine-tuned with LoRA, the metric-depth module stays frozen, and the trajectory predictor and
action decoder are trained with full-parameter updates. Table~\ref{Training_set} lists the
training settings.

\begin{table}[t]
	\centering
	\caption{Detailed training settings.}\label{Training_set}%
	\setlength{\tabcolsep}{12pt}%
	\begin{tabular}{cc}
		\toprule
		\textbf{Setting} & \textbf{Value} \\
		\midrule
        GPUs & 4 $\times$ NVIDIA A800 \\
        Global Batch Size & 64 (16 per device) \\
		Training Steps & 50,000 \\
		Random Seeds & 3 (results averaged) \\
        Optimizer & AdamW \\
        Fine-tuning Method & LoRA \\
		Learning Rate & $2{\times}10^{-4}$ \\
		Warmup Steps & 5,000 \\
		\bottomrule
	\end{tabular}%
\end{table}%

\section{Additional details for simulation experiments}
\label{appendix_c:experiments}

\subsection{Simulation benchmarks and settings}
\label{appendix_c1:simulation_settings}

\begin{figure}[t]
  \centering
  \includegraphics[width=0.65\columnwidth]{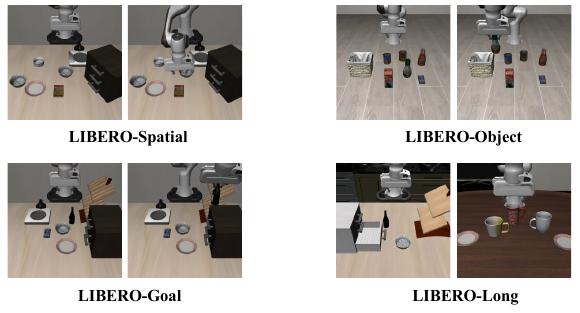}
  \caption{Example scenario from the LIBERO benchmark.}
  \label{fig:libero_figure}
\end{figure}

\paragraph{LIBERO simulation benchmark.} LIBERO comprises four task suites, LIBERO-Spatial,
LIBERO-Object, LIBERO-Goal, and LIBERO-Long, which assess spatial reasoning, object grounding,
goal-directed behavior, and long-horizon planning, respectively. Each suite consists of 10 tasks
with 50 human-teleoperated demonstrations per task. We run 50 rollouts per task
and report the average success rate. Figure~\ref{fig:libero_figure} shows an example scene from each suite, and
Table~\ref{tab:LIBERO_ins} lists example instructions. All ablations in
Section~\ref{sec:ablation_study} run on LIBERO-Long, the suite with the longest tasks.

\begin{table}[t]
\caption{Example language instructions from the four LIBERO suites.}
\label{tab:LIBERO_ins}
\begin{center}
\setlength{\tabcolsep}{4pt}
\begin{tabular}{@{}cc@{}}
		\toprule
		\textbf{LIBERO}  & \textbf{Task Instructions} \\
		\midrule
		Spatial & \makecell[l]{Pick up the black bowl from table center and place it on the
		plate.\\
Pick up the black bowl on the stove and place it on the plate.\\
...}\\

        \midrule
		Object  & \makecell[l]{Pick up the alphabet soup and place it in the basket.\\
Pick up the chocolate pudding and place it in the basket.\\
...}\\
        \midrule
		Goal  & \makecell[l]{Open the top drawer and put the bowl inside.\\
Put the wine bottle on top of the cabinet.\\
Turn on the stove.\\
...}\\
        \midrule
		Long   & \makecell[l]{Put the white mug on the left plate and put the yellow and white
		mug on the right plate.\\
        Put the black bowl in the bottom drawer of the cabinet and close it.\\
        Put the yellow and white mug in the microwave and close it.\\
...}\\
		\bottomrule
	\end{tabular}%
\end{center}
\end{table}

\begin{figure}[t]
  \centering
  \includegraphics[width=0.85\columnwidth]{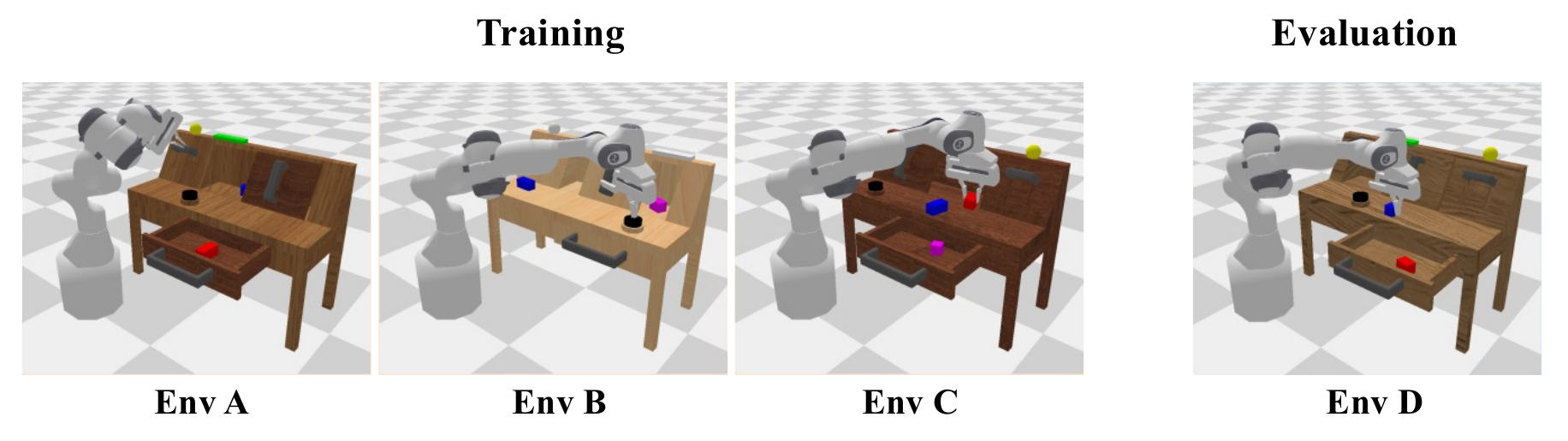}
  \caption{Example scenario from the CALVIN benchmark.}
  \label{fig:calvin_figure}
\end{figure}

\paragraph{CALVIN simulation benchmark.} CALVIN assesses long-horizon, language-conditioned
manipulation in four visually distinct environments, A, B, C, and D, each with a Franka Emika
Panda robot and a tabletop scene with diverse objects, as Figure~\ref{fig:calvin_figure} shows.
Its training data are about 24 hours of teleoperated play, and it defines 34 subtasks, which
Table~\ref{tab:Calvin_ins} groups into eight categories with example instructions. Evaluation
uses 1,000 chains of five consecutive commands executed in order, and we report the success
rate at each chain length and the average sequence length. Under the ABC$\to$D setting of
Section~\ref{sec:simulation_experiments}, policies are trained on environments A, B, and C and
evaluated on environment D, whose scene they never see during training.

\begin{table}[t]
\caption{Example language instructions for the eight CALVIN task categories.}
\label{tab:Calvin_ins}
\begin{center}
\begin{tabular}{cc}
		\toprule
		\textbf{CALVIN}   & \textbf{Task Instructions} \\
		\midrule
        Turn On/Off & Turn on/off lightbulb. Turn on/off LED. \\
        Open/Close Drawer & Open drawer. Close drawer. \\
        Move Slider & Move slider left. Move slider right. \\
        Rotate & Rotate red block left/right. Rotate pink block left/right. ...\\
        Stack/Unstack & Stack block. Unstack block. \\
        Lift & Lift red block table. Lift pink block drawer. ... \\
        Place & Place in slider. Place in drawer. \\
        Push & Push red block left/right. Push into drawer. ...\\
	   \bottomrule
	\end{tabular}%
\end{center}
\end{table}

\subsection{Additional ablation results on LIBERO-Spatial}
\label{appendix_c2:more_ablation}

To check whether the ablation trends on LIBERO-Long hold on another suite,
we repeat the trajectory-predictor, metric-depth, and frame ablations on LIBERO-Spatial,
and Table~\ref{tab:ablation_spatial} reports both suites.
The ranking of supervision targets on LIBERO-Spatial matches that on
LIBERO-Long. The monotonic ladder, from \textit{w/o Traj.} to
\textit{w/ AuxTraj}, \textit{w/ 2D}, \textit{w/ 3D Pos.},
\textit{w/ World}, and OASIS, holds on both suites, with OASIS reaching
99.0\% on Spatial and 95.2\% on Long. The variant \textit{w/ 2D}
stays close to \textit{w/o Traj.} on Long, at
90.7\% versus 89.5\%, and adds a modest gain on Spatial, at 93.3\%
versus 91.6\%, indicating that image-plane geometry alone cannot
align the intermediate with the action space. Moving to \textit{w/ 3D Pos.}, which adds
depth, gains a further 1.6 points on Long and 2.1 on Spatial, and moving to \textit{w/ World},
which adds rotation, gains 0.9 and 1.1 points. OASIS, with camera-frame $SE(3)$, beats \textit{w/ World} by
2.0 points on Long and 2.5 points on Spatial, consistent with deferring
the uncalibrated camera-to-world transform to the learned decoder as
analyzed in Section~\ref{sec:geometric_inconsistency}. Removing metric depth also costs more on Spatial,
5.6 points against 3.4 on Long.

Unlike \textit{w/o Traj.}, \textit{w/ AuxTraj} keeps the trajectory predictor's parameters,
so the OASIS gain of 5.5 points on LIBERO-Long and 7.1 on LIBERO-Spatial over \textit{w/ AuxTraj}
cannot come from trajectory-branch capacity and instead reflects routing the pose-supervised
hidden states to the decoder.

\begin{table}[t]
  \centering
  \caption{Intermediate-type screen on LIBERO-Long and LIBERO-Spatial, plus
  the metric-depth ablation. OASIS, \textit{w/ World}, \textit{w/ 3D Pos.}, and \textit{w/ 2D}
  differ only in the supervision target.
  On both suites, success rises monotonically with its geometric richness.}
  \label{tab:ablation_spatial}
  \begin{tabular}{lcc}
    \toprule
    \textbf{Variant} & \textbf{LIBERO-Long} & \textbf{LIBERO-Spatial} \\
    \midrule
    \textit{w/o Traj.}            & 89.5 & 91.6 \\
    \textit{w/ AuxTraj}           & 89.7 & 91.9 \\
    \textit{w/ 2D}                & 90.7 & 93.3 \\
    \textit{w/o Metric}           & 91.8 & 93.4 \\
    \textit{w/ 3D Pos.}           & 92.3 & 95.4 \\
    \textit{w/ World}             & 93.2 & 96.5 \\
    \oursmidrule
    \textbf{OASIS}                & \textbf{95.2} & \textbf{99.0} \\
    \oursbottomrule
  \end{tabular}
\end{table}

\subsection{Closed-form decoder ablation}
\label{appendix_c3:closed_form_decoder}

To probe whether the learned action decoder is reducible to closed-form geometry,
we replace it with a hardcoded geometric pipeline that converts the same predicted
camera-frame $SE(3)$ trajectory into world-frame poses via known camera extrinsics
and emits relative actions, with gripper open/close further supplied by privileged
simulator information unavailable to OASIS. Despite this strictly stronger setting,
the hardcoded pipeline collapses to 12.4\% on LIBERO-Spatial and 0.0\% on LIBERO-Long
against OASIS's 99.0\% and 95.2\% with the same trajectory predictor, as shown in
Table~\ref{tab:closed_form_decoder}. Rollouts confirm that the predicted trajectory
does encode meaningful geometry, since the robot consistently moves toward the target
object, but small residual prediction noise, contact dynamics, and gripper timing
compound into failures that this closed-form pipeline cannot absorb. This result rules out
the interpretation that OASIS reduces to a renamed closed-form execution pipeline over
predicted poses, and confirms that the learned decoder is necessary.

\begin{table}[t]
  \centering
  \caption{Closed-form decoder ablation. The hardcoded pipeline reads the same
  predicted $SE(3)$ trajectory as OASIS but executes it via known camera extrinsics
  and privileged simulator information for the gripper, a strictly stronger setting
  than OASIS's learned decoder.}
  \label{tab:closed_form_decoder}
  \begin{tabular}{lcc}
    \toprule
    \textbf{Decoder} & \textbf{LIBERO-Spatial} & \textbf{LIBERO-Long} \\
    \midrule
    OASIS (learned decoder)         & \textbf{99.0} & \textbf{95.2} \\
    Hardcoded geometric pipeline    & 12.4          & 0.0           \\
    \bottomrule
  \end{tabular}
\end{table}

\subsection{Decoder attention analysis}
\label{appendix_c4:decoder_attention}

To inspect how the action decoder reads its context, we record the
cross-attention weights of both decoder blocks over 3,372 forward passes
during LIBERO-Long evaluation rollouts with 10 trials per task.
Table~\ref{tab:decoder_attention} averages the first block over its eight
heads, with one row per action step and one column for each of the $H$
trajectory hidden states and for the state embedding
$\mathbf{h}_{\mathrm{state}}$. Every action step places 0.123 to 0.126 of
its attention on a single trajectory hidden state and at most 0.005 on any
other. Per-head inspection shows that one head of the first block gives the
selected state 0.98 to 1.00 of its weight, so the eight-head average is
about $1/8$, while the other seven heads place nearly all their weight on
$\mathbf{h}_{\mathrm{state}}$, which therefore receives about $7/8$ of the
average. Steps 7 and 8 read each other's states, so the mapping is
one-to-one but not index-matched. In the second block, every action step
places an average weight of 1.000 on $\mathbf{h}_{\mathrm{state}}$.

\begin{table}[t]
  \centering
  \caption{Cross-attention of the first decoder block on LIBERO-Long,
  averaged over its eight heads and 3,372 forward passes. Rows are action
  steps, and columns are the eight trajectory hidden states and the state
  embedding. The largest weight on a trajectory hidden state in each row is
  in bold.}
  \label{tab:decoder_attention}
  \small
  \begin{tabular}{cccccccccc}
    \toprule
    & \multicolumn{8}{c}{\textbf{Trajectory hidden state}} & \\
    \cmidrule(lr){2-9}
    \textbf{Step} & 1 & 2 & 3 & 4 & 5 & 6 & 7 & 8 & $\mathbf{h}_{\mathrm{state}}$ \\
    \midrule
    1 & \textbf{0.126} & 0.003 & 0.004 & 0.005 & 0.003 & 0.001 & 0.000 & 0.001 & 0.856 \\
    2 & 0.002 & \textbf{0.123} & 0.001 & 0.000 & 0.000 & 0.000 & 0.001 & 0.001 & 0.872 \\
    3 & 0.000 & 0.000 & \textbf{0.125} & 0.000 & 0.000 & 0.000 & 0.001 & 0.000 & 0.873 \\
    4 & 0.000 & 0.000 & 0.000 & \textbf{0.125} & 0.000 & 0.000 & 0.001 & 0.000 & 0.873 \\
    5 & 0.000 & 0.000 & 0.000 & 0.001 & \textbf{0.124} & 0.001 & 0.000 & 0.000 & 0.873 \\
    6 & 0.000 & 0.000 & 0.000 & 0.000 & 0.000 & \textbf{0.125} & 0.000 & 0.000 & 0.873 \\
    7 & 0.000 & 0.000 & 0.000 & 0.000 & 0.000 & 0.000 & 0.000 & \textbf{0.125} & 0.874 \\
    8 & 0.000 & 0.000 & 0.000 & 0.000 & 0.001 & 0.000 & \textbf{0.125} & 0.000 & 0.873 \\
    \bottomrule
  \end{tabular}
\end{table}

Since the decoder adds no encoding of the order of the trajectory hidden
states, permuting them leaves its output unchanged up to floating-point
summation order. On identical inputs, the \textit{Shuffled-H} permutation
changes no action component by more than
$7.8\times10^{-3}$ in bf16 or $3.3\times10^{-6}$ in fp32, while
\textit{Mean-pool} changes the actions by up to 2.56 on an action scale of
about 2.25.

The attention pattern shows that each action step combines one trajectory
hidden state with the state embedding. The hardcoded pipeline of
Appendix~\ref{appendix_c3:closed_form_decoder} likewise combines a target
pose with the current one, but the two differ in what they consume. The
pipeline reads the decoded trajectory, the six numbers per step
that the linear head produces from $\mathbf{h}_{\mathrm{traj}}$, converts
them with the known camera extrinsics, and executes the resulting geometric
difference. The decoder instead conditions on $\mathbf{h}_{\mathrm{traj}}$
itself and is trained on demonstrated actions rather than made to execute
the plan, so it can act on an imperfect plan instead of executing the
prediction error verbatim.

\subsection{Seed variance of the simulation results}
\label{appendix_c5:seed_variance}

The OASIS results in Tables~\ref{table:libero_benchmark}
and~\ref{table:calvin_benchmark} are means over three random seeds, and
Tables~\ref{tab:seed_variance_libero} and~\ref{tab:seed_variance_calvin} add
their standard deviations. The largest deviation on LIBERO is 0.4 points on Long, where the
three runs reach 95.6\%, 95.2\%, and 94.8\%, so even the lowest run stays
above the 94.0\% of Unified-VLA. On CALVIN, the deviation is at most 0.3
points at every chain length and 0.01 in average sequence length, where even
the lowest run reaches 4.56, above the 4.44 of DreamVLA.

\begin{table}[t]
  \centering
  \caption{Mean and standard deviation of OASIS over three random seeds
  on LIBERO.}
  \label{tab:seed_variance_libero}
  {\small\setlength{\tabcolsep}{5pt}%
  \begin{tabular}{lccccc}
    \toprule
    \textbf{LIBERO} & \textbf{Spatial} & \textbf{Object} & \textbf{Goal} & \textbf{Long} & \textbf{Average} \\
    \midrule
    OASIS & 99.0 $\pm$ 0.2 & 98.8 $\pm$ 0.0 & 97.4 $\pm$ 0.3 & 95.2 $\pm$ 0.4 & 97.6 $\pm$ 0.1 \\
    \bottomrule
  \end{tabular}\par}
  \vspace{4pt}
  \caption{Mean and standard deviation of OASIS over three random seeds
  on CALVIN ABC$\to$D.}
  \label{tab:seed_variance_calvin}
  {\small\setlength{\tabcolsep}{5pt}%
  \begin{tabular}{lcccccc}
    \toprule
    \textbf{CALVIN} & \multicolumn{5}{c}{\textbf{Tasks Completed in a Row}} & \textbf{Avg.} \\
    \cmidrule(lr){2-6}
    & 1 & 2 & 3 & 4 & 5 & \\
    \midrule
    OASIS & 98.1 $\pm$ 0.1 & 94.9 $\pm$ 0.2 & 91.7 $\pm$ 0.2 & 88.9 $\pm$ 0.2 & 83.3 $\pm$ 0.3 & 4.57 $\pm$ 0.01 \\
    \bottomrule
  \end{tabular}\par}
\end{table}

\section{Additional details for real-world experiments}
\label{appendix_d:real_world_experiments}

\subsection{Real-world experimental settings}
\label{appendix_d1:real_world_experiments_settings}

\begin{figure}[tp]
  \centering
  \includegraphics[width=0.95\columnwidth]{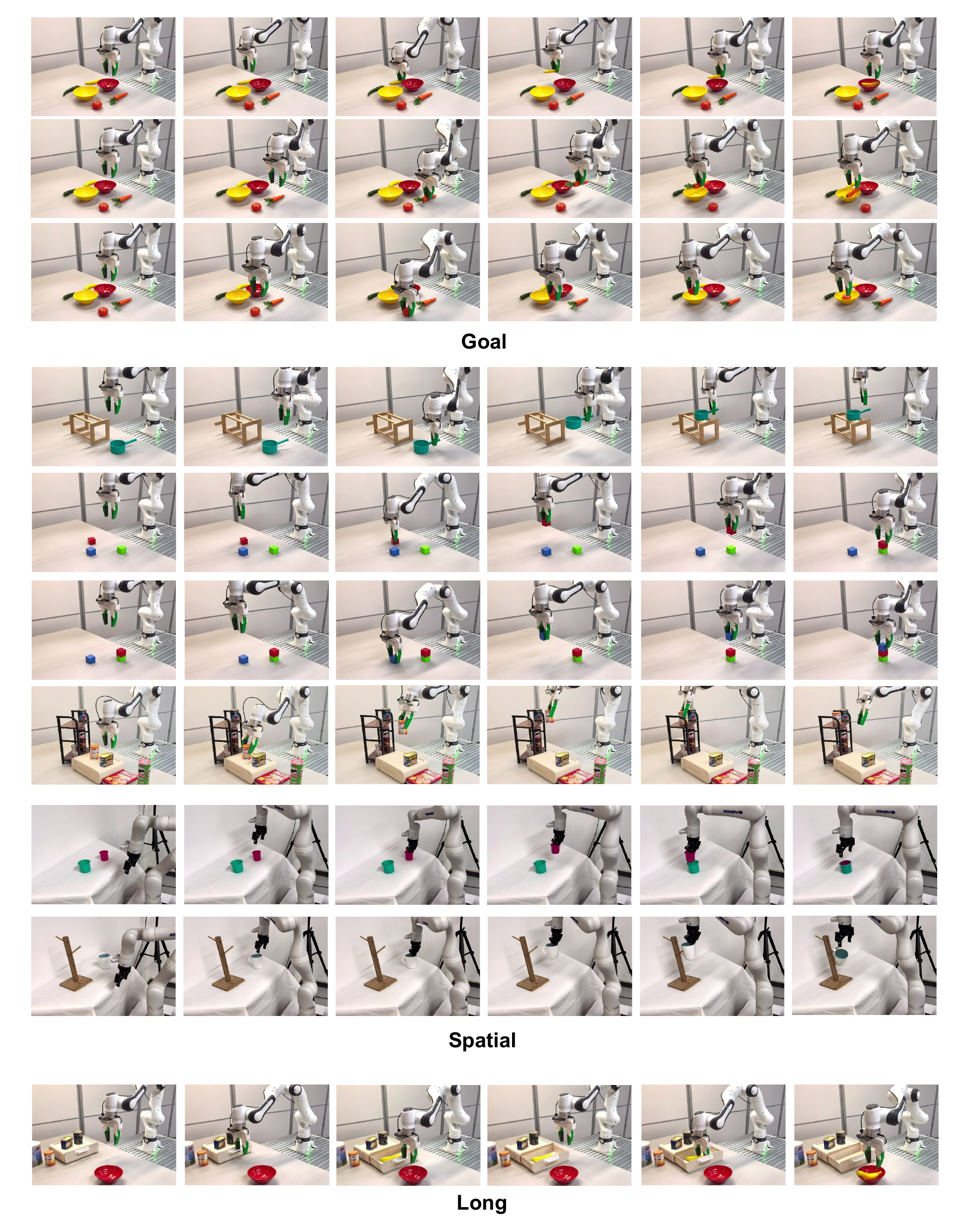}
  \caption{Real-world execution examples of the Goal, Spatial, and Long suites. The Goal
  suite utilizes a tabletop environment with two bowls and four objects arranged in a
  structured layout, where the robot must place the correct object into the correct bowl based
  on the language instruction. The Spatial suite comprises six tasks,
  namely placing the pot on the wooden bracket, stacking blocks, building towers, putting the orange can
  on the shelves, placing the pink cup into the cyan cup, and hanging the cup on the cup holder.
  In the Long task, the robot must execute a multi-stage
  sequence, first opening a drawer, then retrieving the banana from it, and
  finally placing it into the red bowl.}
  \label{fig:realworld_normal}
\end{figure}

The real-world experiments span three suites, with six tasks each in Goal and Spatial and one
in Long. Table~\ref{tab:RealWorld_ins} lists their instructions, and
Figure~\ref{fig:realworld_normal} shows the setup and an execution per suite. Both robots, a
Franka Research 3 and a Kinova Gen3, use a third-person and a wrist camera.

\paragraph{Goal suite.} The robot must place the object named in the instruction into the
named bowl, so the suite tests instruction following across six tasks. The layout of objects
and bowls stays fixed while its distance from the robot base varies across trials, which
prevents memorizing a single configuration and tests robustness to different initial positions.

\begin{table}[t]
\caption{Examples of language instructions for real-world experiments.}
\label{tab:RealWorld_ins}
\begin{center}
\begin{tabular}{cc}
		\toprule
		\textbf{Real-World Experiments}  & \textbf{Task Instructions} \\
		\midrule
		Goal & \makecell[l]{Place the banana into the red bowl.\\
            Pick up the carrot and place it into the red bowl.\\
            Grasp the orange and put it into the yellow bowl.\\
            ...
            }\\

        \midrule
		Spatial  & \makecell[l]{Put the pot on the wooden bracket.\\
            Stack the red cube on top of the green cube.\\
            Place the blue cube on top of the red cube.\\
            Put the orange can on the shelves. \\
            Place the pink cup into the cyan cup.\\
            Hang the cup on the cup holder.\\
            }\\
        \midrule
		Long  & \makecell[l]{
        Open the drawer and place the banana into the red bowl.\\
}\\
		\bottomrule
	\end{tabular}%
\end{center}
\end{table}

\paragraph{Spatial suite.} Its six tasks test spatial reasoning under geometric constraints,
requiring precise localization and placement of objects in various poses. In \textit{Place pot},
both sides of the pot must rest on the two wooden bars, or the pot tips over, and in
\textit{Hang cup}, the handle must engage the cup holder.

\paragraph{Long task.} The Long task chains three stages, opening a drawer, retrieving the banana
from it, and placing the banana into the red bowl, so it demands both precise low-level control
and long-horizon reasoning. Errors compound along the chain, since imprecision while opening the
drawer affects the later grasp, so the policy must correct deviations in closed loop. The
real-world data-scaling study of Appendix~\ref{appendix_d2:real_world_experiments_results} also
uses this task, and Figure~\ref{fig:realworld_failure} shows one of its failures.

\paragraph{Training and evaluation protocol.} All four baselines are rerun under one
protocol shared with OASIS, and Table~\ref{tab:real_world_protocol} lists the
settings of each method. Every method is trained for 50k steps on the same 50
demonstrations per task, with both camera views at $224 \times 224$, and is
evaluated at its final checkpoint over 60 trials per task. No hyperparameter
search or checkpoint selection is performed for any method. On the Goal suite,
the language-conditioned methods are trained jointly on all tasks and ACT is
trained per task, while on the Spatial and Long suites every method is trained
per task. Since $\pi_{0.5}$, RDT, and Seer-Large start from robotic-pretrained
weights while ACT and OASIS do not, the robotic-pretraining axis favors the
baselines. Under this protocol, OASIS still leads $\pi_{0.5}$, the strongest
baseline, by 7.6 points on average, 89.2\% versus 81.6\% in
Table~\ref{table:multitask_results}.

\begin{table}[t]
  \centering
  \caption{Initialization, learning rate, and batch size of each method in the
  real-world comparison. Each baseline keeps the default learning rate of its
  official implementation, and batch sizes are given as the batch per GPU times
  the number of GPUs.}
  \label{tab:real_world_protocol}
  \begin{tabular}{llcc}
    \toprule
    \textbf{Method} & \textbf{Initialization} & \textbf{Learning rate} & \textbf{Batch size} \\
    \midrule
    $\pi_{0.5}$ & Official pretrained     & $5{\times}10^{-5}$ & $16 \times 4$ \\
    RDT         & Official pretrained     & $1{\times}10^{-4}$ & $16 \times 4$ \\
    Seer-Large  & Official pretrained     & $1{\times}10^{-4}$ & $16 \times 4$ \\
    ACT         & From scratch            & $1{\times}10^{-4}$ & $16 \times 4$ \\
    OASIS       & No robotic pretraining  & $2{\times}10^{-4}$ & $16 \times 4$ \\
    \bottomrule
  \end{tabular}
\end{table}

\begin{figure}[t]
  \centering
  \includegraphics[width=0.95\columnwidth]{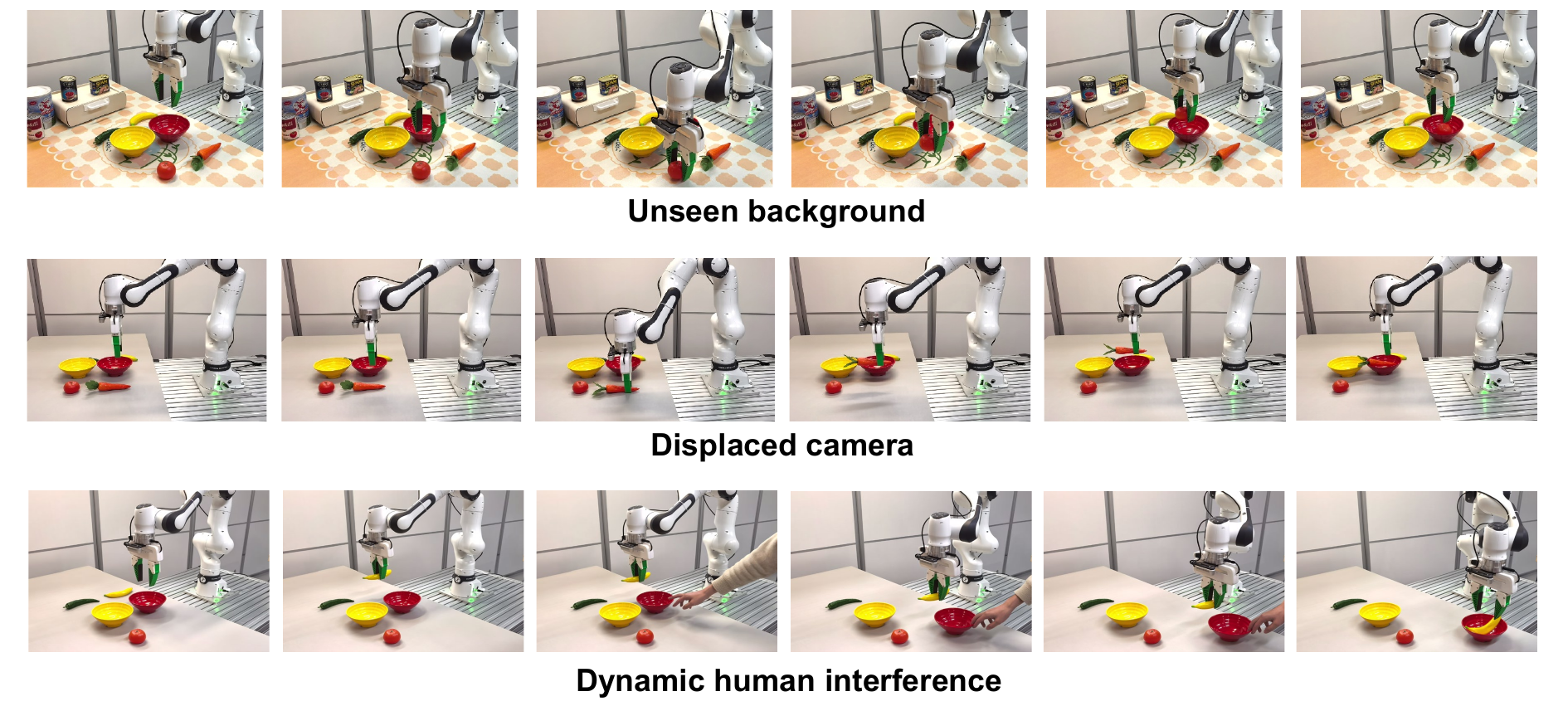}
  \caption{Real-world execution examples under the three OOD scenarios.}
  \label{fig:realworld_ood}
\end{figure}

\paragraph{OOD scenario settings.} To evaluate zero-shot generalization, we build three
out-of-distribution scenarios on the Goal suite, illustrated in Figure~\ref{fig:realworld_ood}.
The unseen backgrounds scenario adds objects and a tabletop that never appear in training,
testing whether a policy attends to task-relevant features and ignores visual distractors.
The displaced camera scenario moves only the third-person camera at test
time, by a translation of about 15\,cm
relative to the training viewpoint, while the wrist camera and all other components
remain unchanged. No recalibration or test-time adaptation is performed for any policy.
This configuration probes robustness to a single uncalibrated third-person viewpoint
shift. The human interference scenario relocates the target bowl while
the robot is transporting the object, testing whether a policy can retarget in closed loop
when its goal moves. Figure~\ref{fig:ood_results} in Section~\ref{sec:real_world_experiments}
reports the resulting success rates of OASIS, $\pi_{0.5}$, and RDT, and shows that OASIS loses less
than $\pi_{0.5}$ under each of the three scenarios.
Against the in-distribution Goal success, the drops are 3.8, 7.4, and 12.1 points for OASIS
and 6.4, 18.6, and 13.5 for $\pi_{0.5}$.

\subsection{Additional real-world experimental results}
\label{appendix_d2:real_world_experiments_results}

\begin{figure}[t]
  \centering
  \includegraphics[width=0.95\columnwidth]{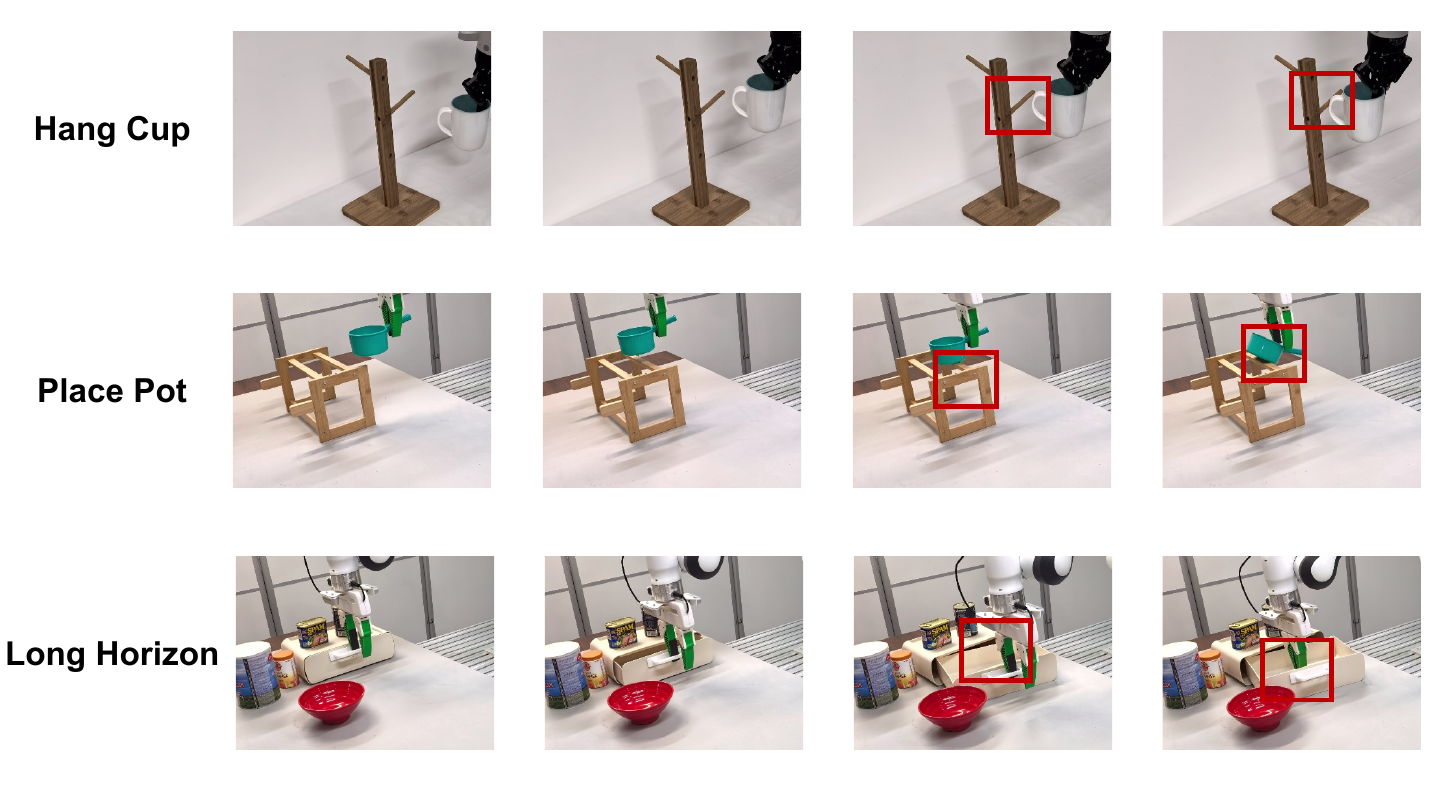}
  \caption{Failure cases of OASIS in real-world execution.}
  \label{fig:realworld_failure}
\end{figure}

\paragraph{Failure analysis.} As shown in Figure~\ref{fig:realworld_failure} and Table~\ref{tab:RealWorld_result},
three of the four lowest real-world success rates involve tasks that demand
sub-centimeter placement or precise orientation. \textit{Hang cup}, at 76.6\%,
fails when a small orientation deviation keeps the handle from engaging the
holder, and \textit{Place pot}, at 83.3\%, fails when small x-axis or yaw deviations
leave the pot resting on a single supporting bar. The \textit{Long horizon}
task, at 83.3\%, is dominated by stage compounding: imprecision during
drawer opening shifts the banana's reachable region and propagates into
grasp failures downstream. Figure~\ref{fig:realworld_failure} shows one failure of each of these three
tasks, with the failure marked by a red box.

To test whether the \textit{Hang cup} and \textit{Place pot} failures stem
from trajectory prediction, Table~\ref{tab:failure_error} measures the
position and rotation errors of the predicted trajectory offline on
in-distribution demonstrations from both platforms, with separate columns for
the last quarter of each trajectory, where fine manipulation happens. Among
the tasks in the table, the one with the lowest success rate on each
platform has the most accurate predicted trajectory in every column, and the
same ordering holds for median errors. Only the ordering within a platform is
informative, since rotation errors differ several-fold across platforms.
Prediction accuracy therefore does not appear to be the main bottleneck, and
the failures point instead to execution tolerance. The pot, for example,
must rest on both wooden bars, so deviations that the other tasks absorb are
enough to fail it.

\begin{table}[t]
\caption{Per-task success rates of OASIS on the real-world Goal, Spatial, and Long suites.}
\label{tab:RealWorld_result}
\begin{center}
\begin{tabular}{clc}
    \toprule
    \textbf{Category} & \textbf{Task Description} & \textbf{Success Rate (\%)} \\
    \midrule
    \multirow{6}{*}{Goal}
      & Place the banana into the red bowl & 100.0 \\
      & Place the banana into the yellow bowl & 100.0 \\
      & Place the carrot into the red bowl & 100.0 \\
      & Place the carrot into the yellow bowl & 100.0 \\
      & Place the orange into the red bowl & 95.0 \\
      & Place the orange into the yellow bowl & 96.7 \\
    \midrule
    \multirow{6}{*}{Spatial}
      & Stack blocks & 90.0 \\
      & Build towers & 90.0 \\
      & Place pot on wooden bracket & 83.3 \\
      & Put orange can on shelves & 81.6 \\
      & Hang the cup & 76.6 \\
      & Place the pink cup & 93.3 \\
    \midrule
    Long
      & Open the drawer and place the banana into the red bowl & 83.3 \\
    \bottomrule
\end{tabular}
\end{center}
\end{table}

\begin{table}[t]
  \centering
  \caption{Trajectory prediction error on in-distribution demonstrations,
  measured offline. SR is the success rate in percent, and Pos. and Rot. are
  the mean position error in cm and the mean rotation error in degrees. The
  most accurate task on each platform is in bold.}
  \label{tab:failure_error}
  \renewcommand{\arraystretch}{1.5}
  \setlength{\tabcolsep}{10pt}
  \begin{tabular}{llccccc}
    \toprule
    & & & \multicolumn{2}{c}{\textbf{Full trajectory}} & \multicolumn{2}{c}{\textbf{Last 25\%}} \\
    \cmidrule(lr){4-5} \cmidrule(lr){6-7}
    \textbf{Platform} & \textbf{Task} & \textbf{SR} & \textbf{Pos.} & \textbf{Rot.} & \textbf{Pos.} & \textbf{Rot.} \\
    \midrule
    \multirow{2}{*}{Kinova} & Hang the cup & 76.6 & \textbf{1.02} & \textbf{0.08} & \textbf{1.33} & \textbf{0.11} \\
     & Place the pink cup & 93.3 & 1.27 & 0.09 & 1.97 & 0.16 \\
    \midrule
    \multirow{3}{*}{Franka} & Place pot on wooden bracket & 83.3 & \textbf{0.83} & \textbf{0.33} & \textbf{1.25} & \textbf{0.32} \\
     & Stack blocks & 90.0 & 1.02 & 0.38 & 1.69 & 0.47 \\
     & Build towers & 90.0 & 0.91 & 0.68 & 1.31 & 0.95 \\
    \bottomrule
  \end{tabular}
\end{table}

\paragraph{Inference speed analysis.} To deploy the trained policy without operating-system
or code-version conflicts, we wrap model loading and inference in a WebSocket server on a
remote machine. The server communicates in real time with the robot's control computer,
receiving live camera images at up to 30 Hz and the end-effector state, and returning
actions for execution. On an NVIDIA RTX 4090 GPU without additional computational load,
the policy consumes about 4.5 GB of GPU memory. Each inference cycle predicts the $SE(3)$
trajectory and 8 subsequent action steps, executed sequentially. The end-to-end response
time of this server loop, including the WebSocket transport, is about 0.05 s,
corresponding to roughly 20 Hz. To isolate policy inference, we also time one
inference on a single RTX 4090 with the transport removed, at batch size 1 in bf16, taking
the mean over 100 runs after 10 warm-up iterations. OASIS takes 45 ms, against 120 ms for
$\pi_{0.5}$ and 260 ms for RDT, so it runs 2.7 and 5.8 times faster on the same GPU. Each
number is the time to produce one action chunk, so for the flow-matching and diffusion
baselines it includes their sampling iterations.

\paragraph{Data-scaling on the real-world Long task.} To probe how data efficiency
interacts with the aligned-intermediate design, we train OASIS and $\pi_{0.5}$ on
the real-world Long task with progressively reduced demonstration budgets and
evaluate each variant over three independent runs of 20 trials each, for 60 trials per
configuration. Table~\ref{tab:realworld_scaling} reports
success rate as a function of demonstration count. With only 10 demonstrations,
OASIS reaches 35.0\% while $\pi_{0.5}$ reaches 15.0\%. At 25 demonstrations, OASIS
reaches 55.0\% while $\pi_{0.5}$ reaches 35.0\%, and at the full 50 demonstrations,
OASIS reaches 83.3\% while $\pi_{0.5}$ reaches 71.6\%. OASIS thus matches the
observed 25-demonstration success rate of $\pi_{0.5}$ using only 10 demonstrations
on this task.

\begin{table}[t]
\caption{Data-scaling results on the real-world Long task. We train OASIS and
$\pi_{0.5}$ with 10, 25, and the full 50 demonstrations and evaluate each variant
over three independent runs of 20 trials each, for 60 trials per configuration. All
numbers are success rates in percent.}
\label{tab:realworld_scaling}
\begin{center}
\begin{tabular}{lccc}
    \toprule
    \textbf{Method} & \textbf{10 demos} & \textbf{25 demos} & \textbf{50 demos} \\
    \midrule
    $\pi_{0.5}$           & 15.0 & 35.0 & 71.6 \\
    \textbf{OASIS (Ours)} & \textbf{35.0} & \textbf{55.0} & \textbf{83.3} \\
    \bottomrule
\end{tabular}
\end{center}
\end{table}

\subsection{Wilson 95\% confidence intervals for real-world results}
\label{appendix_d3:wilson_cis}

To quantify finite-trial uncertainty, we report Wilson score 95\% confidence
intervals in Table~\ref{tab:wilson_main} for the head-to-head OASIS versus $\pi_{0.5}$ comparison on the
real-world main results in Table~\ref{table:multitask_results}. Goal and
Spatial each comprise six tasks evaluated over 60 trials per task,
so we report a pooled descriptive Wilson interval over $n{=}360$ trials
per suite, while Long is a single task with $n{=}60$.
The intervals overlap by under one point on Goal and Spatial and widely on Long.

\begin{table}[t]
\caption{Wilson 95\% confidence intervals, in percent, for OASIS vs $\pi_{0.5}$
on the real-world results.}
\label{tab:wilson_main}
\begin{center}
\begin{tabular}{lccc}
    \toprule
    \textbf{Method} & \textbf{Goal} & \textbf{Spatial} & \textbf{Long} \\
    \midrule
    $\pi_{0.5}$           & 95.0 [92.2, 96.8] & 78.3 [73.8, 82.3] & 71.6 [59.2, 81.5] \\
    \textbf{OASIS (Ours)} & 98.6 [96.8, 99.4] & 85.8 [81.9, 89.1] & 83.3 [72.0, 90.7] \\
    \bottomrule
\end{tabular}
\end{center}
\end{table}

\section{Broader impacts}
\label{appendix_e:broader_impacts}

OASIS proposes a new design principle for visuomotor policies, aligning the
intermediate representation with the action space via $SE(3)$ end-effector
trajectory prediction. Instead of decoding actions from observation-space
features alone, our framework first predicts a camera-frame $SE(3)$ trajectory
whose pose-supervised hidden states condition the action decoder. This aligned
intermediate improves manipulation precision and generalization across
simulation benchmarks and real-world platforms.

A key strength of OASIS lies in its training efficiency and accessibility. It
requires neither large-scale robotic pretraining nor annotated spatial labels,
with supervision derived solely from standard expert demonstrations. With only
0.18B trainable parameters, trained on four GPUs and run on a single RTX 4090,
the framework lowers the cost barrier for research groups and small organizations
developing manipulation policies. Separating geometric prediction from action
execution also makes the intermediate representation inspectable, which can
support debugging and safer deployment.

Practically, this design can benefit assistive household robots, industrial
manipulators, and tabletop laboratory automation, where precise $SE(3)$ control
under modest data budgets is essential. We hope OASIS inspires further work
on geometrically aligned intermediates and low-cost robot learning.

OASIS also carries risks. A learned policy can still fail under
conditions outside its training data, and our OOD tests show success dropping by up
to 12.1 points. A robot acting on a wrong prediction near
people could break objects or cause injury, so deployment outside the lab calls for
speed and force limits, a monitored workspace, and a human who can stop the robot at
any time.

\end{document}